\documentclass[twoside,11pt]{article}

\usepackage{blindtext}

\usepackage{jmlr2e}

\usepackage{lastpage}

\usepackage{setspace}
\usepackage{graphicx,subcaption}
\usepackage{xcolor}
\hypersetup{
	colorlinks,
	linkcolor={red!50!black},
	citecolor={blue!50!black},
	urlcolor={blue!80!black}
}

\usepackage{amsmath}
\usepackage{amssymb}
\usepackage{bbm}
\usepackage{systeme}

\usepackage[linesnumbered,lined,boxed,ruled]{algorithm2e}

\SetCommentSty{mycommfont}
\SetKwInput{KwInput}{Input}                
\SetKwInput{KwOutput}{Output}              

\ShortHeadings{Graph Construction using Principal Axis Trees for Simple Graph Convolution}{Graph Construction using Principal Axis Trees for Simple Graph Convolution}
\firstpageno{1}

\begin{document}

\title{Graph Construction using Principal Axis Trees for Simple Graph Convolution}

\author{\name Mashaan Alshammari \email mashaan.awad1930@alum.kfupm.edu.sa \\
       \addr Independent Researcher\\
       Riyadh, Saudi Arabia
       \AND
       \name John Stavrakakis \email john.stavrakakis@sydney.edu.au \\
       \addr School of Computer Science\\
       The University of Sydney\\
       NSW 2006, Australia
       \AND
       \name Adel F. Ahmed \email adelahmed@kfupm.edu.sa \\
       \addr Information and Computer Science Department\\
       King Fahd University of Petroleum and Minerals\\
       Dhahran, Saudi Arabia
       \AND
       \name Masahiro Takatsuka \email masa.takatsuka@sydney.edu.au \\
       \addr School of Computer Science\\
       The University of Sydney\\
       NSW 2006, Australia}

\editor{}

\maketitle

\begin{abstract}
	Graph Neural Networks (GNNs) are increasingly becoming the favorite method for graph learning. They exploit the semi-supervised nature of deep learning, and they bypass computational bottlenecks associated with traditional graph learning methods. In addition to the feature matrix $X$, GNNs need an adjacency matrix $A$ to perform feature propagation. In many cases, the adjacency matrix $A$ is missing. We introduce a graph construction scheme that constructs the adjacency matrix $A$ using unsupervised and supervised information. Unsupervised information characterizes the neighborhood around points. We used Principal Axis trees (PA-trees) as a source for unsupervised information, where we create edges between points falling onto the same leaf node. For supervised information, we used the concept of penalty and intrinsic graphs. A penalty graph connects points with different class labels, whereas an intrinsic graph connects points with the same class labels. We used the penalty and intrinsic graphs to remove or add edges to the graph constructed via PA-tree. We tested this graph construction scheme on two well-known GNNs: 1) Graph Convolutional Network (GCN) and 2) Simple Graph Convolution (SGC). The experiments show that it is better to use SGC because it is faster and delivers better or the same results as GCN. We also test the effect of oversmoothing on both GCN and SGC. We found out that the level of smoothing has to be carefully selected for SGC to avoid oversmoothing.
\end{abstract}


\section{Introduction}\label{Introduction}

Graph representation learning methods have gained popularity in recent years. The reason was the simplicity of modelling most of machine learning problems using graph representation. Given a set of samples $X$ with $\{\vec{x_1}, \vec{x_2}, \cdots, \vec{x_n} \}$, one could construct a graph $G(V,E)$, where the set $V$ contains the feature vectors as graph nodes. The set $E$ holds the relations between feature vectors represented as graph edges \cite{hart2000pattern}. Graph Neural Network (GNNs) is one of the most effective schemes for graph representation learning. It has been applied to sentiment analysis \cite{ZHOU2020Modeling,LIANG2021dependency,PHAN2022Convolutional} and computer vision tasks \cite{QI2021Deep,Zhang2022Graph,JIA2022Data}.

The idea of designing a deep network for graph representation learning has come through multiple iterations \cite{ Hammond2011Wavelets,Bruna2013Spectral,Henaff2015Deep,Defferrard2016Convolutional}. One of the well-known GNN methods is Graph Convolutional Network (GCN) \cite{kipf2017semi}. Throughout the hidden layers, GCN performs feature propagation between neighbors on the graph, then a nonlinear transformation of the graph is passed to the next layer. The last layer in GCN was set as a softmax function to produce the labels on graph nodes. Simple Graph Convolution (SGC) was proposed by \cite{Wu2019Simplifying} where they remove the nonlinearity between the layers. This means stacking $K$ hidden layers is a matrix multiplication between the feature matrix $X$ and the adjacency matrix $A$ for $K$ times.

GCN and SGC both cannot create or modify graph edges, which means the graph has to be constructed before running GCN or SGC. There are methods to modify the adjacency matrix $A$ to achieve higher accuracy \cite{franceschi2019learning} or to improve robustness against attacks \cite{jin2020graph}. The problem is, these methods only work with GCN, because they need the transition between the hidden layers to perform adjacency matrix optimization. Since this transition is absent in SGC and replaced by matrix multiplication, these adjacency optimization methods are not compatible with SGC.

We propose a new graph construction scheme based on unsupervised and supervised information. The proposed scheme works with GCN and SGC. For unsupervised graph edges creation, we used Principal Axis trees (PA-trees) \cite{sproull1991refinements,McNames2001Fast}, which is one type of Binary Space Partitioning trees (BSP-trees) \cite{Ram2013Which}. We also used supervised information to create graph edges. The field of dimensionality reduction introduced the concept of penalty and intrinsic graph. We used this concept to create edges from the training data \cite{Yan2007Graph}. Both unsupervised and supervised information were blended in one adjacency matrix and processed by either GCN or SGC.

Our contributions can be summarized as the following:

\begin{itemize}
	\item The proposed graph construction scheme uses Principal Axis trees (PA-trees) to highlight the density in the dataset. It also utilizes the training data to characterize intraclass compactness and interclass separability.
	
	\item We studied the effect of smoothing on GCN and SGC using the proposed graph construction. Our results provide an empirical evidence that SGC is more vulnerable to oversmoothing than GCN. This supports the findings presented in \cite{Zhao2019PairNorm,Yang2020Revisiting}.
\end{itemize}

\section{Related work}
\label{RelatedWork}

The learning task on graphs consists of two components: graph construction and learning algorithm. This study focuses on learning algorithms using deep learning. The next subsection introduces Graph Convolutional Network (GCN), while the following subsection discusses graph construction methods.

\subsection{Graph Convolutional Network (GCN)}
\label{section:GCN}
Performing learning tasks on graphs is one of the long-studied problems in machine learning literature. One of the oldest methods in this field is spectral clustering \cite{ Shi1997Normalized,Weiss1999Segmentation,Shi2000Normalized,Ng2001Spectral,Luxburg2007tutorial}. Given a set of samples $X$ with $\{\vec{x_1}, \vec{x_2}, \cdots, \vec{x_n} \}$, spectral clustering method starts by constructing the adjacency matrix $A \in \mathbb{R}^{n\times n}$ using some pairwise similarity metric and the degree matrix $D_{ii}=\sum_{j} A_{ij}$. Then, eigen decomposition is performed on the graph Laplacian $L=D^{-1/2}AD^{-1/2}$ to map the points into an embedding space. In that space, similar points fall closer to each other and can be detected using $k$-means. The biggest hurdle for spectral clustering is decomposing an $n \times n$ adjacency matrix, which can be prohibitive with large datasets \cite{Defferrard2016Convolutional,Shaham2018SpectralNet}.

The superiority of spectral clustering comes from the mapping function. So, the question in the literature was: can we learn this mapping function instead of computing it through eigen decomposition. Designing a deep network that learns new representations of the feature vectors $\{\vec{x_1}, \vec{x_2}, \cdots, \vec{x_n} \}$, can replace the deterministic mapping function in spectral clustering.

The connectivity of the graph $G$ is encoded in the graph Laplacian $L$, and the graph Laplacian eigenvectors define the graph Fourier transform \cite{Shuman2013emerging}. Finding the new representation of a feature vector $x$ is done by performing the convolution between the input signal $x$ with a filter $g \in \mathbb{R}^n$ \cite{Wu2021Comprehensive}:

\begin{equation}
	x *_G g = U(U^\top x \odot U^\top g),
	\label{Eq-GraphConv-1}
\end{equation}
\noindent
where $U$ is the matrix of the eigenvectors of the graph Laplacian and $\odot$ is the elementwise product. If we set the filter $g_\theta$ as $diag(U^\top g)$, then the formula in equation \ref{Eq-GraphConv-1} can be rewritten as:

\begin{equation}
	x *_G g_\theta = U g_\theta U^\top x.
	\label{Eq-GraphConv-2}
\end{equation}

One of the earliest studies in this field was done by \cite{Bruna2013Spectral}, where they designed a convolutional net that operates on the spectrum of the input features. 
However, their approach still needs the expensive eigen decomposition step with complexity $O(n^3)$. The new direction of research was to approximate the filter $g_{\theta}$ using Chebyshev polynomials. This idea has came through a series of refinements by different studies \cite{Henaff2015Deep,Defferrard2016Convolutional}. Graph convolutional network (GCN) introduces a first order approximation of Chebyshev polynomials. Approximating graph convolutions using Chebyshev polynomials takes the following form:

\begin{equation}
	x *_G g_\theta = \sum_{i=0}^{K} \theta_i T_i(\tilde{L})x,
	\label{Eq-GCN-1}
\end{equation}
\noindent
where $\tilde{L}=\frac{2L}{\lambda_{max}}-I_n$. The Chebyshev polynomials are defined recursively as: $T_i(x)=2xT_{i-1}(x)-T_{i-2}(x)$ with $T_{0}(x)=1$ and $T_{1}(x)=x$. GCN assumes that $K=1$ and $\lambda_{max}=2$ \cite{kipf2017semi}, therefore equation \ref{Eq-GCN-1} is simplified as:

\begin{equation}
	x *_G g_\theta = \theta_0 x - \theta_1 D^{-1/2}AD^{-1/2} x.
	\label{Eq-GCN-2}
\end{equation}
\noindent
GCN further assumes that $\theta = \theta_0 = -\theta_1$, which leads to a simpler definition of graph convolution: 

\begin{equation}
	x *_G g_\theta = \theta(I_n + D^{-1/2}AD^{-1/2})x.
	\label{Eq-GCN-3}
\end{equation}
\noindent
GCN is a multilayer network where a single layer is defined as:
\begin{equation}
	H = X *_G g_\Theta = f(\bar{A}X\Theta),
	\label{Eq-GCN-4}
\end{equation}
\noindent
where $X$ is the feature matrix holding the feature vectors $\{\vec{x_1}, \vec{x_2}, \cdots, \vec{x_n} \}$, and $\bar{A}=\tilde{D}^{-1/2}\tilde{A}\tilde{D}^{-1/2}$ with self loops added to the adjacency $\tilde{A} = A+I_n$. $f(\cdot)$ is the activation function which was set as $ReLU(x)=max(0,x)$ for the hidden layers and $softmax(x)$ for the last layer.

A modification to GCN was introduced as Simple Graph Convolution (SGC) \cite{Wu2019Simplifying}. SGC removes the nonlinearity between GCN layers. In SGC, the learned representations $\hat{Y}$ of the input feature vectors $X$ is defined as:
\begin{equation}
	\hat{Y} = softmax(\bar{A}\cdots \bar{A}\bar{A} X \Theta^{(1)}\Theta^{(2)} \cdots \Theta^{(K)}),
	\label{Eq-SGC-1}
\end{equation}
\noindent
where $K$ is the number of layers. Let $\bar{A}^K$ denote the repeated multiplication of the adjacency matrix and $\Theta = \Theta^{(1)}\Theta^{(2)} \cdots \Theta^{(K)}$. Then, equation \ref{Eq-SGC-1} can be rewritten as:
\begin{equation}
	\hat{Y} = softmax(\bar{A}^K X \Theta).
	\label{Eq-SGC-2}
\end{equation}
Given this definition, SGC brought down the computations in the hidden layers to a pre-processing step with no weights needed $\bar{X} = \bar{A}^K X$. The final layer becomes a linear logistic regression classifier $\hat{Y} = softmax(\bar{X}\Theta)$.

Graph Convolutional Network (GCN) is still an active research area with several topics emerging from the literature. One of the topics is Graph AutoEncoder (GAE) where GCN is employed to compute note representations in the latent space \cite{kipf2016variational}. A new contribution to this research was to replace the weight sharing in GCNs with factor sharing between reconstructed adjacency matrices to find similarities \cite{Chen2023Dual}. Another research track studies adversarial attacks on GCNs \cite{Dai2018Adversarial}. Recent work by \cite{Wu2022Robust} proposed multi-view graph augmentation to defend against adversarial attacks.

\subsection{Graph construction}
\label{section:GraphConstruction}

The methods introduced in the previous section need an adjacency matrix $A$ to work on. For many applications, the adjacency matrix $A$ is not present, and researchers have to construct it from the feature matrix $X$. It is important to mention some studies that perform adjacency matrix modifications while training the GCN. For example, \cite{franceschi2019learning} designed a framework that modifies the adjacency matrix to improve the performance of GCN. \cite{jin2020graph} modified the adjacency matrix to prevent malicious attacks from compromising the learning algorithm. Some researchers relied on feedback from GCN to improve the adjacency. \cite{Zhong2023Contrastive} used a self-adaptive adjacency matrix network that learns the adjacency matrix based on feedback from the GCN network. The GCN network performs pseudo-labeling on the data. Then, the adjacency matrix learns the connections adaptively. They constructed the initial graph using the $k$-nn graph. These methods use alternating optimization schema to update $\theta$ and $A$. They are incompatible with SGC because SGC computes the hidden layers as a pre-processing step with no weight optimization.

Another approach is to construct the adjacency matrix $A$ independently from GCN. \cite{ye2021coupled} used Gaussian kernel to compute pairwise similarities between samples. The Gaussian kernel is a conventional choice to construct the adjacency matrix, but it needs the setting of the hyperparameter $\sigma$. Another option is to involve supervised information in the adjacency matrix construction. \cite{ma2023homophily} proposed a new metric named cross-class neighborhood similarity (CCNS) to measure the similarity between nodes. CCNS quantifies how similar the neighborhoods of two nodes with the same label are across the entire graph.

Constructing the adjacency matrix $A$ involves identifying similar points and creating an edge linking them. In its simplest form, the adjacency matrix $A$ can be constructed using the Gaussian kernel:
\begin{equation}
	A_{ij}=\exp{\left(\frac{-d^2\left(i,j\right)}{\sigma}\right)},
	\label{Eq-HeatKernel}
\end{equation}
\noindent
where $-d^2\left(i,j\right)$ is the Euclidean distance, and $\sigma$ is a global scale set manually. Points separated by a small Euclidean distance are linked by an edge with large weight. There are two problems associated with constructing the adjacency matrix using the formula in equation \ref{Eq-HeatKernel}: 1) the tuning of the parameter $\sigma$ and 2) the resulting adjacency matrix is not sparse.

Binary space-partitioning trees (BSP-trees) are very useful to define a hierarchical structure of the dataset \cite{Ram2013Which}. As we go deeper down a BSP-tree, the relevant neighborhood around the point $x$ is narrowed down. One of the famous BSP-trees is the principal-axis tree (PA-tree) \cite{sproull1991refinements, McNames2001Fast}. A PA-tree splits the feature vectors at the median along the first principal component. Constructing the adjacency matrix $A$ from PA-tree can be done by creating edges linking the points that fall into the same leaf node. Defining similarity between points using binary space-partitioning trees was implemented for spectral clustering \cite{Yan2019Similarity,Wang2019DC2}.

Graph construction using binary space-partitioning trees (BSP-trees) is done in an unsupervised way. Studies in the field of dimensionality reduction have constructed the graph using supervised information. A study by \cite{Yan2007Graph} proposed the concept of penalty and intrinsic graphs. Edges in the penalty graph $G^p$ connect points from different classes. These edges were used to characterize the interclass separability. For intraclass compactness, they used the intrinsic graph $G^i$ that connects points from the same class.

From the review introduced in this section, we can identify three conditions for the graph $G$ to be passed to a deep network. First, it has to work with both GCN and SGC, regardless of the fact that SGC skips the nonlinearity between the hidden layers. Second, most of the graph edges have to be constructed in an unsupervised manner. Finally, the construction scheme must use the training samples to add edges to the graph or remove edges that link samples from different classes.

\section{Graph construction for GCN and SGC}
\label{ProposedApproach}

Our proposed graph construction method passes through two stages: 1) constructing graph edges using unsupervised information and 2) adding/removing edges from the graph based on supervised information. The proposed model expects feature vectors with the same dimensions. In the case of feature vectors with different dimensions, a preprocessing step is needed to ensure equal dimensions are passed to the neural net. The preprocessing can take the form of feature selection where the features with the most importance are kept. Another option is to apply dimensionality reduction on a subset of the features. The next subsections introduce the problem statement followed by graph construction stages.

\subsection{Problem statement}
\label{ProblemStatement}

The task of the proposed method is to perform node classification on the graph using two types of Graph Neural Networks (GNNs): GCN and SGC. There are some notations to be introduced before we present the problem statement. Let $G=(V,E)$ be a graph where $V$ is the set of nodes and $E$ is the set of edges. Graph edges describe the similarity between each pair of points and represented by the adjacency matrix $A \in \mathbb{R}^{n \times n}$. The feature matrix $X=\{\vec{x_1}, \vec{x_2}, \cdots, \vec{x_n} \}\in \mathbb{R}^{n \times d}$, where $x_i$ is the feature vector for the node $v_i$. The graph $G$ can be represented using the adjacency and feature matrices $G=(A,X)$. In a node classification problem, only a subset of nodes $V_l=\{v_1, v_2, \cdots, v_l \}$ have known class labels $Y_l=\{y_1, y_2, \cdots, y_l \}$. The goal for a GNN is to learn a function $f_\theta: V_l \rightarrow Y_l$ that maps nodes to their corresponding labels, then it can uncover the labels for unseen data.

With the introduction of these notations, the problem can be stated as follows:

\noindent
\textit{Given a feature matrix $X$ and partial node label $Y_l$ in absence of the adjacency matrix $A$, construct $A$ using a PA-tree, add/remove edges using penalty graph $G^p$ and intrinsic graph $G^i$, then run GNN to perform node classification.}

\subsection{Constructing a graph using unsupervised information from PA-trees}
\label{GraphPA-trees}

Binary Space Partitioning trees (BSP-trees) provide a hierarchical view for the input points. Principal Axis trees (PA-trees) are one type of BSP-trees. The PA-tree algorithm starts by projecting all points in the dataset onto the first principal component, and split them at the median. Points that are less than the median are placed into the left child and other points are placed into the right child. This process is repeated recursively until a maximum number of data points in leaf node $n_0$ is reached \cite{Keivani2021Random}. Algorithm \ref{Alg:ConstructPATree} shows the steps for PA-tree construction.

\begin{figure}
	\begin{algorithm}[H]
		\DontPrintSemicolon
		
		\KwInput{feature matrix $X$, maximum number of points in a leaf node $n_0$}
		\KwOutput{tree data structure}
		
		\SetKwProg{Fn}{Function}{:}{}
 		\SetKwFunction{FTree}{MakeTree}
 		\SetKwFunction{FRule}{Rule}
 		
  		\Fn{\FTree{$S$, $n_0$}}
  		{
 			\eIf {$ \left| S \right| \le n_0$}
 			{ 				
				\KwRet leaf containing $S$
 			} 			
 			{
	 			Set $u$ to be the first principal component in $S$
	 			
	 			Let $c$ to be the median of projection of $S$ onto $u$
	 			
				\FRule($x$) = $ \left( x^\top U \le c \right) $
				
				LSTree = \FTree{$x \in S:$ \FRule($x$) = true, $n_0$}
				
				RSTree = \FTree{$x \in S:$ \FRule($x$) = false, $n_0$}
				
				\KwRet LSTree, RSTree
			}
 		}
		
		\caption{PA-tree construction}
		\label{Alg:ConstructPATree}
	\end{algorithm}
\end{figure}

We created edges from PA-trees by connecting the points falling into the same leaf node:

\begin{equation}
	(x_i,x_j) \in E^{PA} \Leftrightarrow x_i \in W \ and \ x_j \in W,
	\label{Eq-PA-tree-1}
\end{equation}
\noindent
where $W$ is a leaf node. One parameter that influences this process is $n_0$, which is the maximum number of points allowed in a leaf node to stop splitting. In the experiments we set $n_0=20$, the same setting was used by \cite{Yan2018Nearest,Yan2021Nearest}.

\begin{figure*}
	\centering
	\begin{subfigure}[t]{0.40\textwidth}
		\centering\includegraphics[width=\textwidth,keepaspectratio]{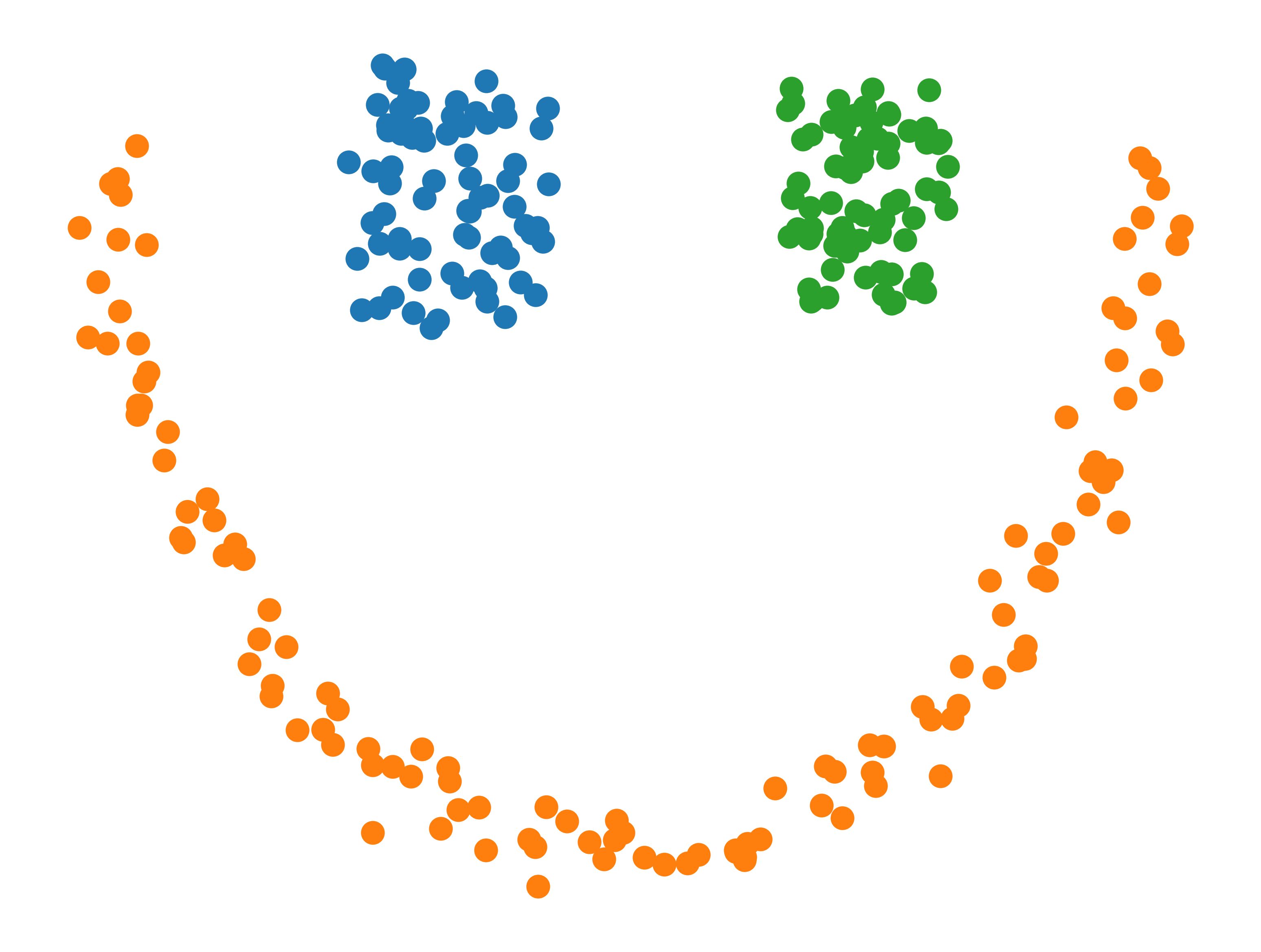}
		\caption{input features}
	\end{subfigure}
	\begin{subfigure}[t]{0.40\textwidth}
		\centering\includegraphics[width=\textwidth,keepaspectratio]{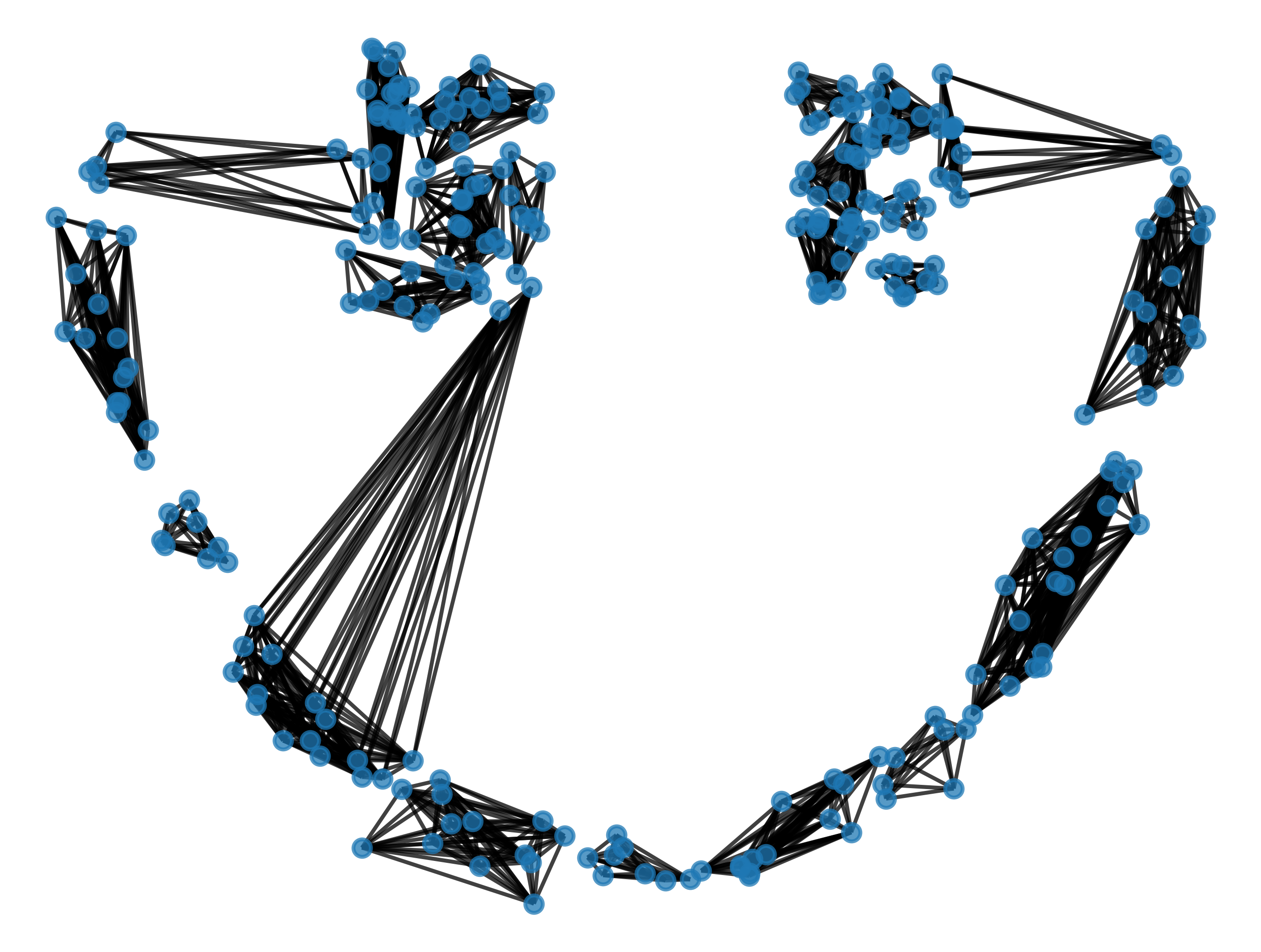}
		\caption{PA-tree graph $G^{PA}$}
	\end{subfigure}
	
	\caption{A graph constructed using principal Axis trees (PA-trees); (a) original feature matrix with three classes; (b) the graph obtained by connecting the points falling onto the same leaf node in a PA-tree. (Best viewed in color)}
	\label{Fig:GraphPA-trees}
\end{figure*}

Figure\ \ref{Fig:GraphPA-trees} shows an example of constructing a graph using PA-trees. All points falling onto the same leaf node were fully connected. The points in the orange class that represents the smile, were split into two different tree branches. This can be explained by the position of this class. It stretches along the first principal component, and splitting at the median will break this class. This observation shows the importance of using supervised information to fill in these gaps created by unsupervised construction of the graph.

\subsection{Constructing penalty and intrinsic graphs from the training data}
\label{GraphPenalty}

\begin{figure*}
	\centering
	\begin{subfigure}[t]{0.32\textwidth}
		\centering\includegraphics[width=\textwidth]{features.png}
		\caption{input features}
	\end{subfigure}
	\begin{subfigure}[t]{0.32\textwidth}
		\centering\includegraphics[width=\textwidth]{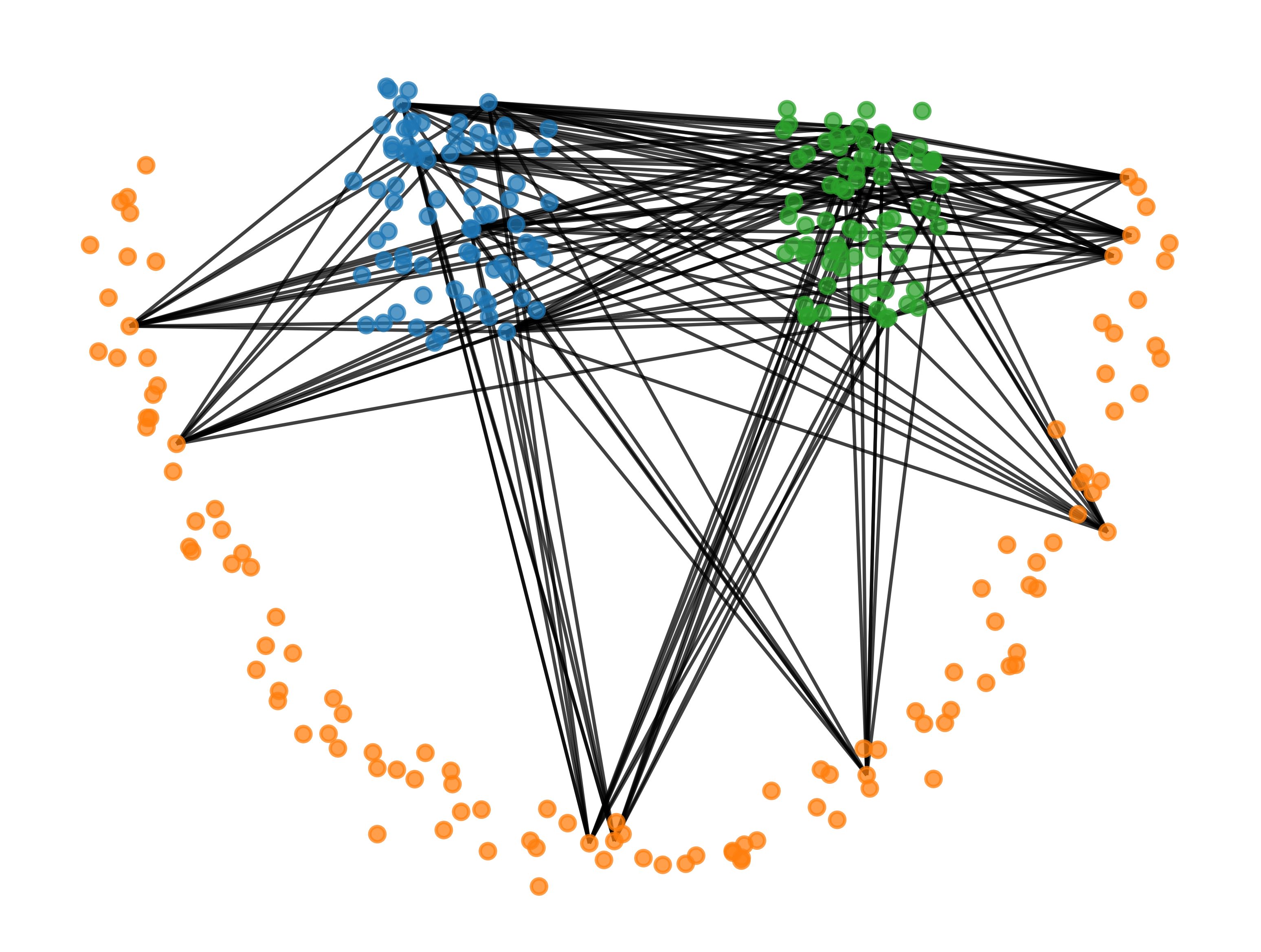}
		\caption{penalty graph $G^p$}
	\end{subfigure}
	\begin{subfigure}[t]{0.32\textwidth}
		\centering\includegraphics[width=\textwidth]{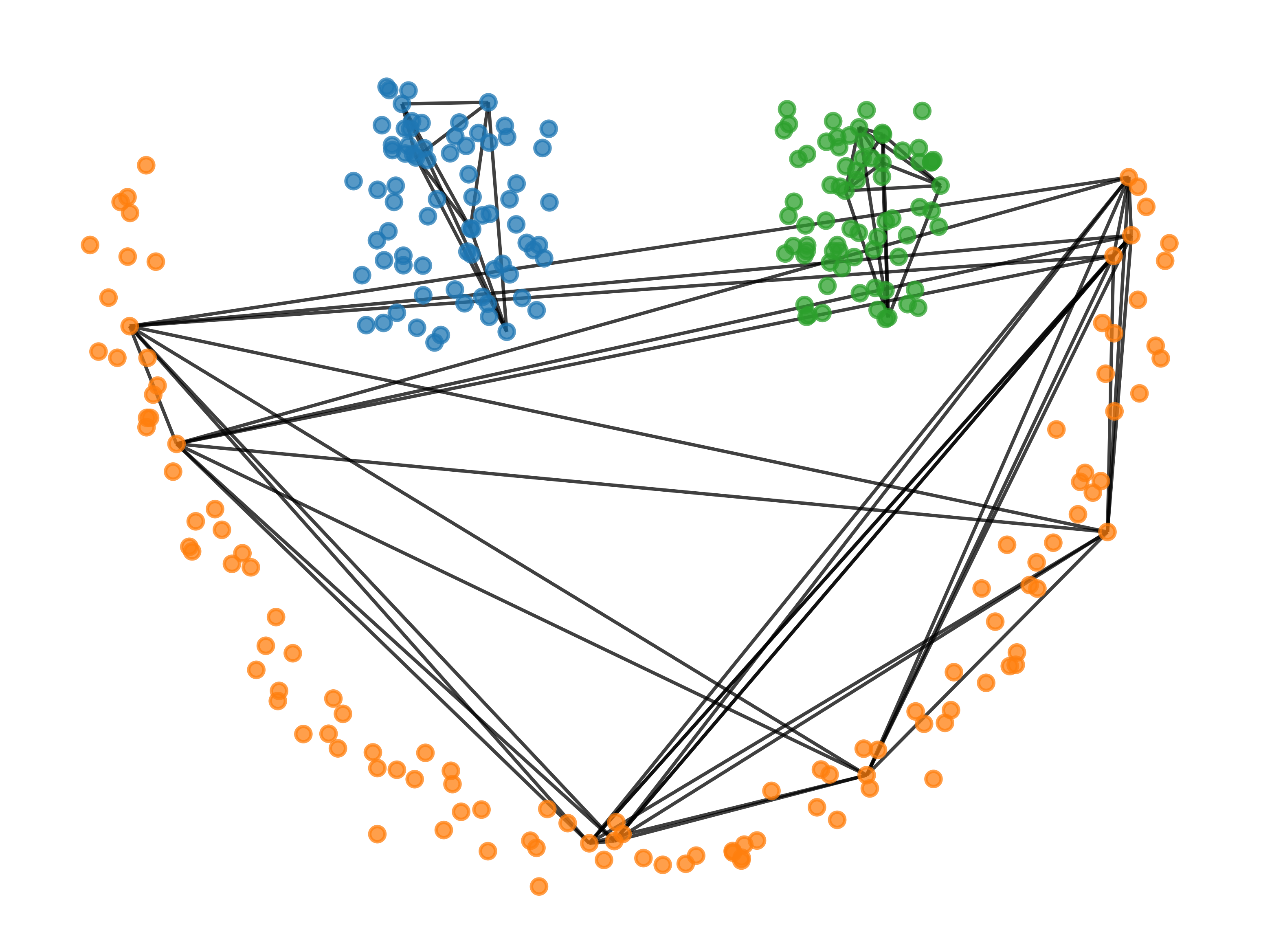}
		\caption{intrinsic graph $G^i$}
	\end{subfigure}
	
	\caption{Constructing penalty and intrinsic graphs using training data; (a) original feature matrix with three classes; (b) the penalty graph connecting samples from different classes; (c) the intrinsic graph connecting samples from the same class. (Best viewed in color)}
	\label{Fig:Graph-penalty}
\end{figure*}

From the graph shown in Figure\ \ref{Fig:GraphPA-trees}, it is evident that unsupervised information cannot capture the high-level relationships between classes. We have to use the training feature vectors to capture these high-level relationships between classes. \cite{Yan2007Graph} presented a framework to construct edges from the training feature vectors. They constructed two graphs, a penalty graph $G^p$ (Figure\ \ref{Fig:Graph-penalty}-b) with edges connecting samples from different classes. This graph characterizes the interclass separability and defined as:

\begin{equation}
	(x_i,x_j) \in E^{p} \Leftrightarrow y_i \ne y_j,
	\label{Eq-PenaltyGraph-1}
\end{equation}
\noindent
where $y_i$ and $y_j$ are the class labels for the feature vectors $x_i$ and $x_j$ respectively.

The second graph is the intrinsic graph $G^i$ (Figure\ \ref{Fig:Graph-penalty}-c) which connects samples from the same class to identify the intraclass compactness. The intrinsic graph $G^i$ is defined as:

\begin{equation}
	(x_i,x_j) \in E^{i} \Leftrightarrow y_i = y_j.
	\label{Eq-IntrinsicGraph-1}
\end{equation}

The proposed method relies on the unsupervised edges made by the PA-tree. The graph created these edges based on the first principal components, which means it will split the points along the axis with the highest variance. However, this is not necessarily the case with all classes. Some classes stretch along the first principal component, for example, the class with the orange color in Figure\ \ref{Fig:GraphPA-trees}-b. The intrinsic graph helps us to compensate for such weaknesses. The intrinsic graph creates edges based on the class not the location of a point. For example, the class with the orange color was connected by the intrinsic graph in Figure\ \ref{Fig:Graph-penalty}-c.

The final adjacency matrix that was passed to GCN and SGC is the result of refining the graph produced by the PA-tree $G^{PA}$. All edges in the penalty graph $G^p$ should be removed from the PA-tree graph $G^{PA}$. Also, all edges in the intrinsic graph $G^i$ are added to the PA-tree graph $G^{PA}$. The pseudocode in Algorithm \ref{Alg:ConstructGraph} shows the steps for our graph construction method.

\begin{figure}
	\begin{algorithm}[H]
		\DontPrintSemicolon
		
		\KwInput{feature matrix $X$, partial labels $Y_l$}
		\KwOutput{adjacency matrix $A$}
		
		Construct a PA-tree using the feature matrix $X$
		
		Construct a PA-tree graph: $(x_i,x_j) \in E^{PA} \Leftrightarrow x_i \in W \ and \ x_j \in W$
		
		Construct a penalty graph using $Y_l$: $(x_i,x_j) \in E^{p} \Leftrightarrow y_i \ne y_j$
		
		Construct an intrinsic graph using $Y_l$: $(x_i,x_j) \in E^{i} \Leftrightarrow y_i = y_j$
		
		Add edges in the intrinsic graph $G^i$ to the PA-tree graph $G^{PA}$:
		$A = A^{PA} + A^i$
		
		Remove edges in the penalty graph $G^p$ from the final graph:		
		$A = A - A^p$		
		
		\caption{Graph construction for GCN and SGC}
		\label{Alg:ConstructGraph}
	\end{algorithm}
\end{figure}

\section{Experiments and discussions}
\label{Experiments}

We designed the experiments to test the efficiency of the proposed graph construction scheme. We also examined different settings that have an influence over the learning algorithm. These settings include the level of smoothing used in GCN and SGC and the number of trees used to construct the graph. Our experiments include a comparison with ground truth adjacency and machine learning methods. 

All the datasets used in the experiments are available publicly. We downloaded some datasets from the scikit-learn library \cite{scikit-learn,sklearn_api}, and we downloaded others from public repositories. Table \ref{Table:Datasets-Properties} shows the properties of the datasets, their training splits, and their sources. We used three groups, each of which has four datasets. The first group is the 2-dimensional datasets \texttt{Dataset 1} to \texttt{Dataset 4}. These are easy to visualize with artificially created classes to make it harder for the classifier. The second group of datasets involves \texttt{iris}, \texttt{wine}, \texttt{BC-Wisc.}, and \texttt{digits}. These datasets have a large number of dimensions. The last group of datasets involves \texttt{Olivetti}, \texttt{PenDigits}, \texttt{mGamma}, and \texttt{credit card}. These datasets have a large number of nodes, which will be useful in testing the running time for both SGC and GCN.

All experiments were coded in Python 3, and can be found on the following GitHub repository \url{https://github.com/mashaan14/PAtree-SGC}. Here are the properties of the machine used in the experiments: a Windows 11 machine with 20 GB of memory and a 3.10 GHz Intel Core i5-10500 CPU.

\begin{table}
	\centering
	\caption{Properties of tested datasets. $N =$ number of samples, $d =$ number of features.}
	\begin{tabular}{l|r|r|c|c} 
		\hline
        & $N$ & $d$ & Train/Valid/Test & source \\
		\hline		
		\texttt{Dataset 1}& 266 & 2 & 50/50/166& \cite{Zelnik2005Self}\\		
        \texttt{Dataset 2}& 399& 2& 50/50/199&\cite{ClusteringDatasets}\\
        \texttt{Dataset 3}& 622& 2& 50/50/522&\cite{Zelnik2005Self}\\
        \texttt{Dataset 4}& 788& 2& 50/50/688&\cite{ClusteringDatasets}\\
        \hline
        \texttt{iris}& 150& 4& 50/50/50&\cite{scikit-learn}\\
        \texttt{wine}& 178& 13& 50/50/78&\cite{scikit-learn}\\
        \texttt{BC-Wisc.}& 569& 30& 50/50/469&\cite{scikit-learn}\\
        \texttt{digits}& 1797& 64& 50/50/1697&\cite{scikit-learn}\\
        \hline
        \texttt{Olivetti}& 400& 4096& 1024/1024/2048&\cite{scikit-learn}\\
        \texttt{PenDigits}& 10992& 16& 2748/2748/5496&\cite{Dua2019UCI}\\
        \texttt{mGamma}& 19020& 10& 4755/4755/9510&\cite{Dua2019UCI}\\
        \texttt{credit card}& 30000& 24& 7500/7500/15000&\cite{Dua2019UCI}\\
        \hline
	\end{tabular}
	\label{Table:Datasets-Properties}
\end{table}

\begin{figure}
	\centering
	\begin{subfigure}[t]{0.24\textwidth}
		\centering\includegraphics[width=\textwidth]{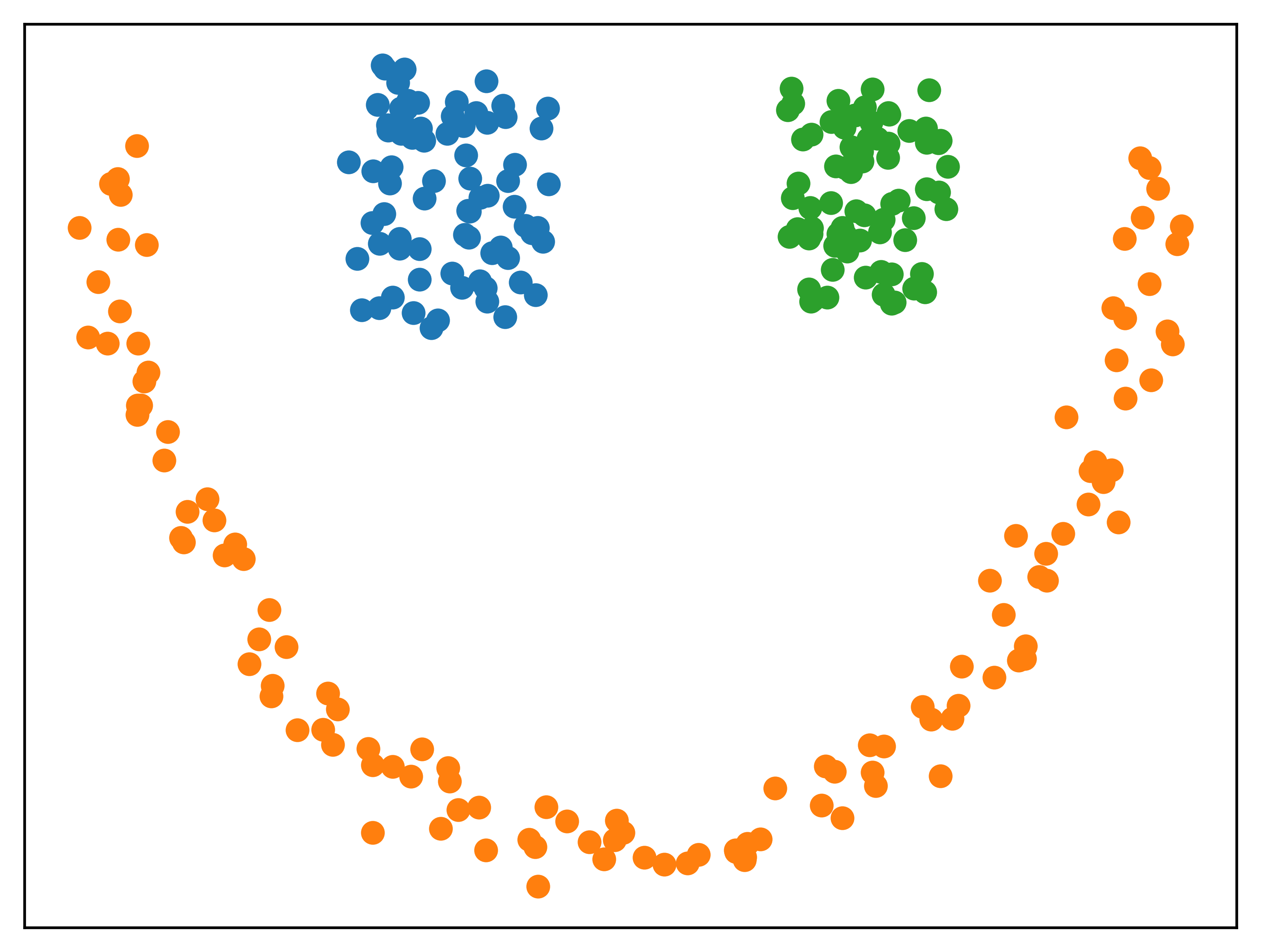}
		\caption{\texttt{Dataset 1}}
	\end{subfigure}
	\begin{subfigure}[t]{0.24\textwidth}
		\centering\includegraphics[width=\textwidth]{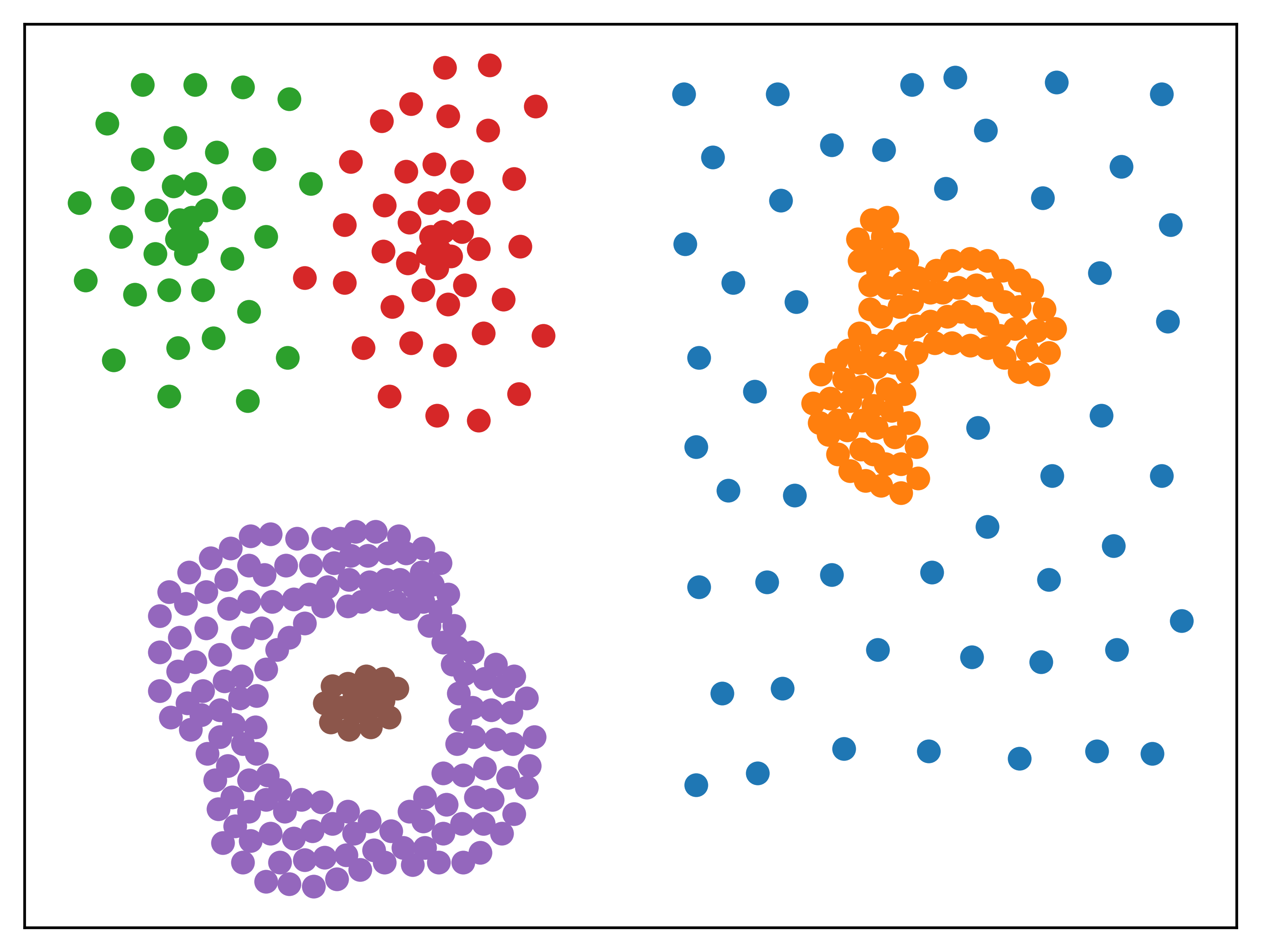}
		\caption{\texttt{Dataset 2}}
	\end{subfigure}
	\begin{subfigure}[t]{0.24\textwidth}
		\centering\includegraphics[width=\textwidth]{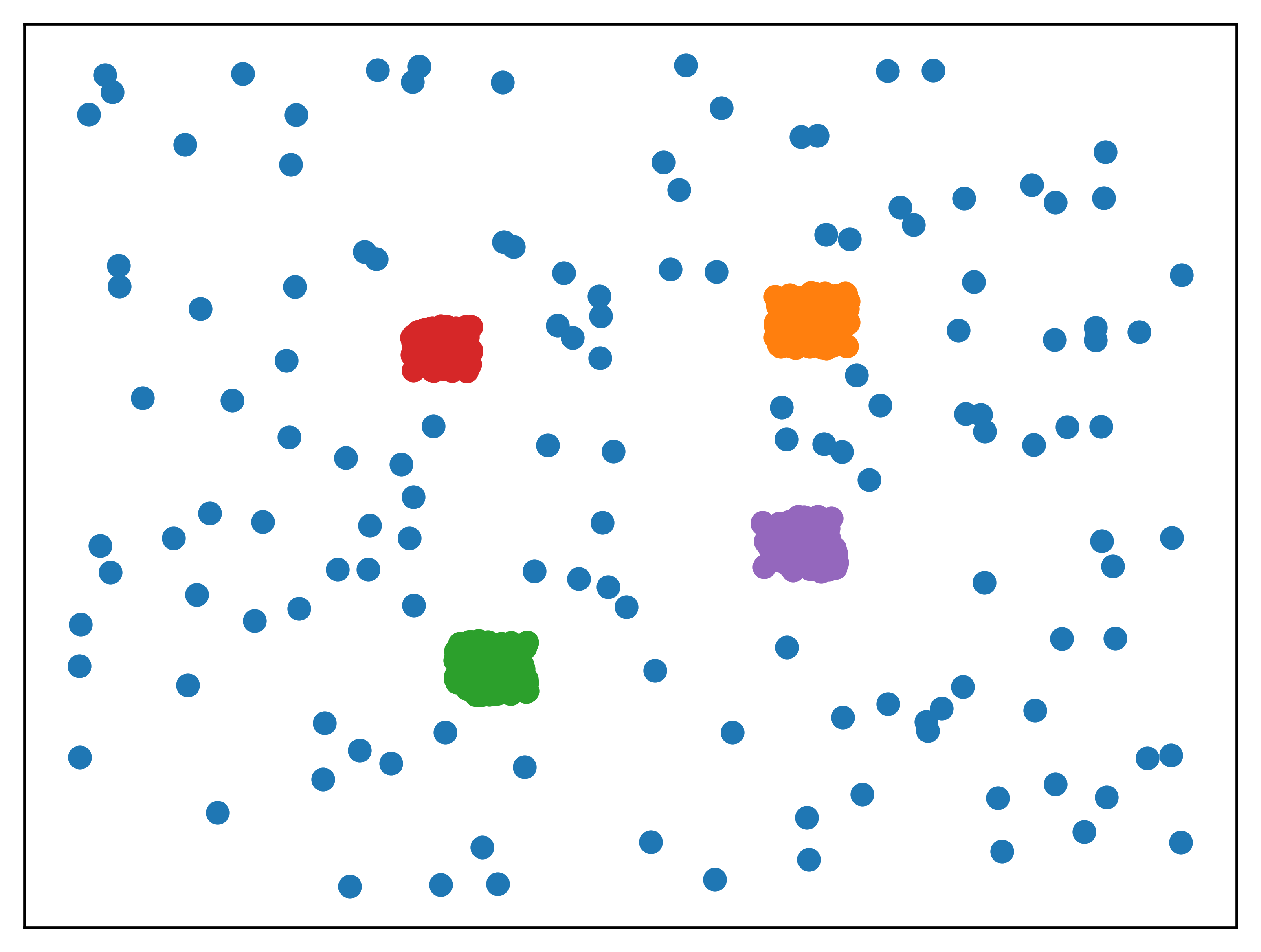}
		\caption{\texttt{Dataset 3}}
	\end{subfigure}
	\begin{subfigure}[t]{0.24\textwidth}
		\centering\includegraphics[width=\textwidth]{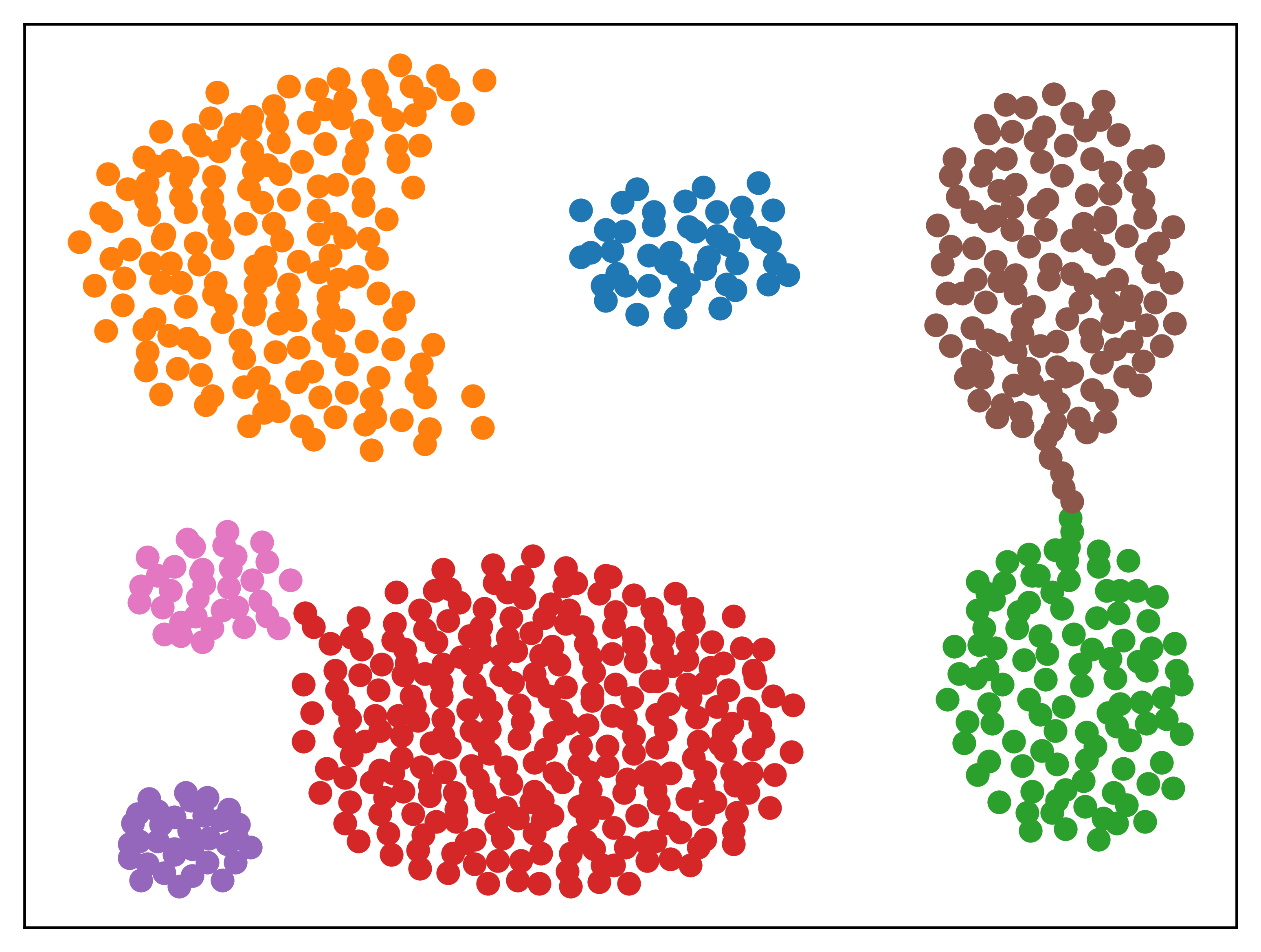}
		\caption{\texttt{Dataset 4}}
	\end{subfigure}
	
	\caption{small-size datasets used in the experiments; from left to right \texttt{Dataset 1} to \texttt{Dataset 4}; source: \cite{Zelnik2005Self,ClusteringDatasets}. (Best viewed in color)}
	\label{Fig:small-datasets}
\end{figure}

\subsection{Testing the accuracy of GCN and SGC}
\label{Experiments-Accuracy}

The results for testing the accuracy of GCN and SGC are shown in Table\ \ref{Table:Ex-Accuracy}. With \texttt{Dataset 2} and \texttt{Dataset 4}, SGC outperforms GCN. But SGC falls short in \texttt{Dataset 1} and \texttt{Dataset 3}. This can be explained by the nature of the sparse classes (i.e., points within the same class do not share a single mean). Since SGC performs feature propagation in the original feature space, points in these sparse classes got pulled towards other classes’ means. GCN can be very useful in these situations because it uses nonlinear transformation after each round of feature propagation. These nonlinear transformations map the points in sparse classes closer to each other, which helps improving the accuracy.

\begin{table}
	\centering
	\includegraphics[width=0.6\textwidth]{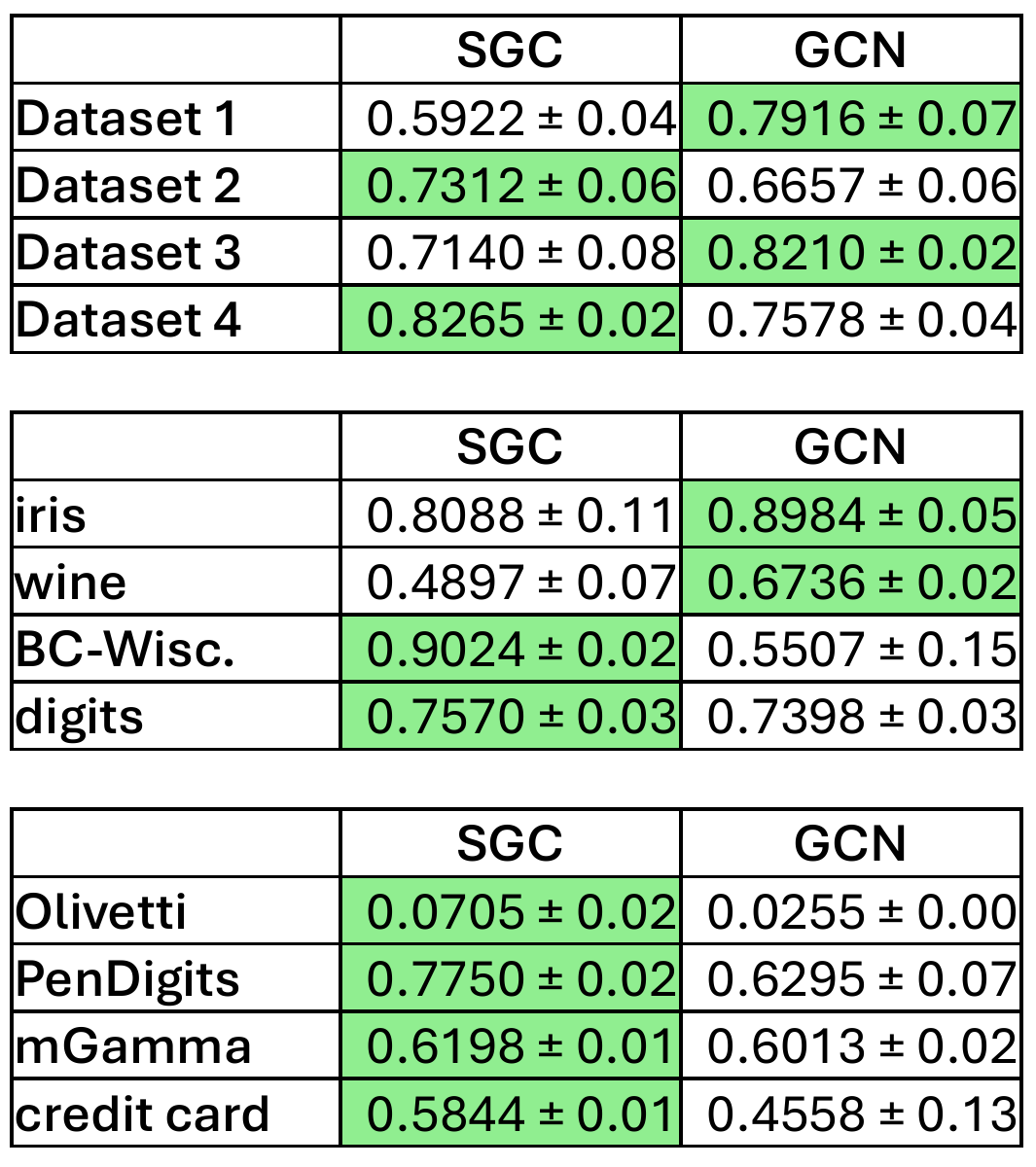}
	\caption{Accuracy scores for SGC and GCN using different datasets; the top two tables are averages for 50 runs; the bottom table contains averages for 10 runs; the best scores are highlighted in green. (Best viewed in color)}
	\label{Table:Ex-Accuracy}
\end{table}

With the last eight datasets, SGC outperforms GCN in six datasets out of eight. This performance by SGC can be explained by the fact that most classes in these datasets have a single mean. Unlike small datasets, which are usually designed artificially with sparse classes, classes in real dataset have their own mean. GCN was the best performer in \texttt{iris} and \texttt{wine} datasets. These datasets contain sparse classes, which can be better separated by nonlinear transformation.

\begin{figure*}
	\centering
	\begin{subfigure}[t]{0.32\textwidth}
		\centering\includegraphics[width=\textwidth]{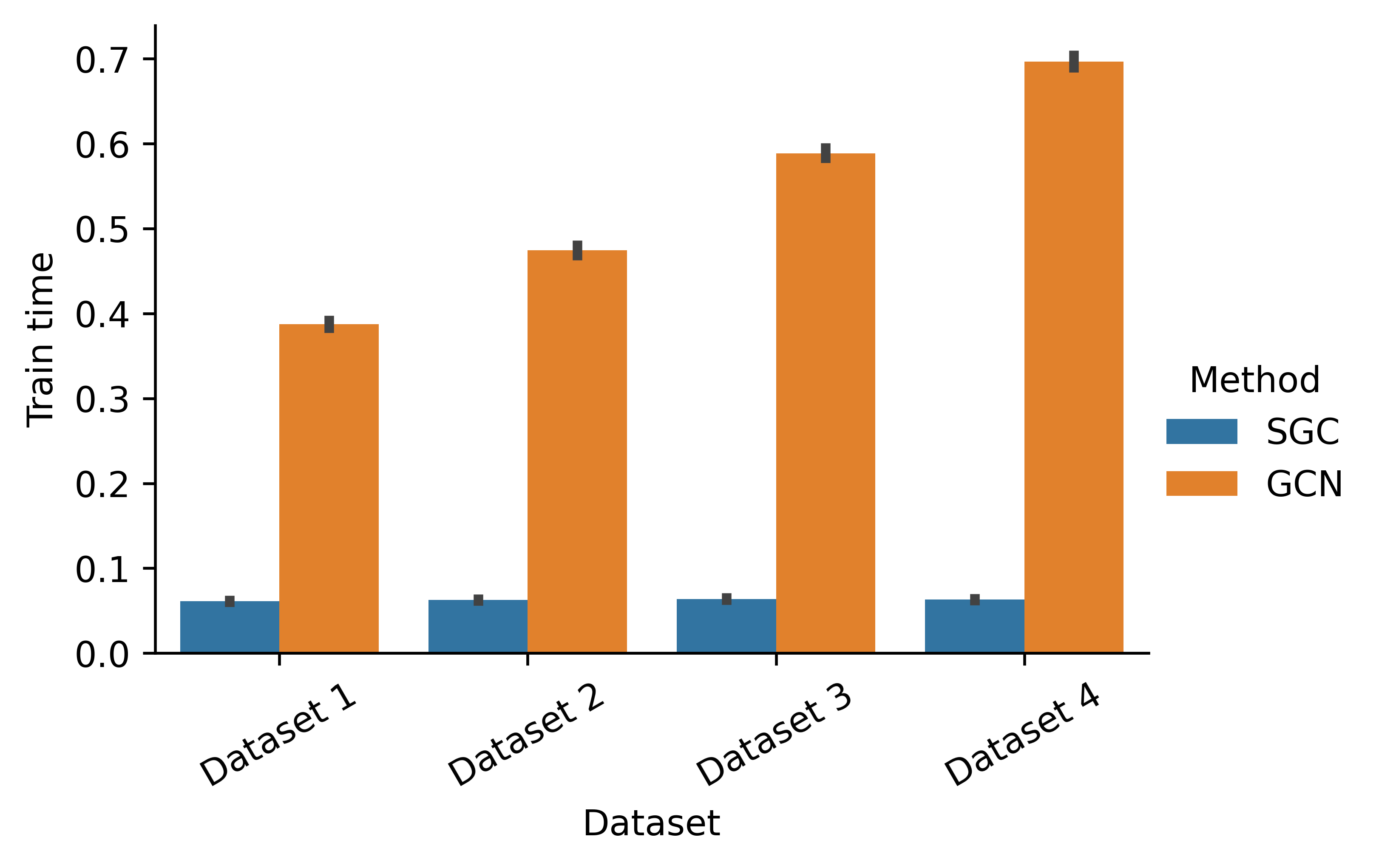}
		\caption{small-size datasets}
	\end{subfigure}
	\begin{subfigure}[t]{0.32\textwidth}
		\centering\includegraphics[width=\textwidth]{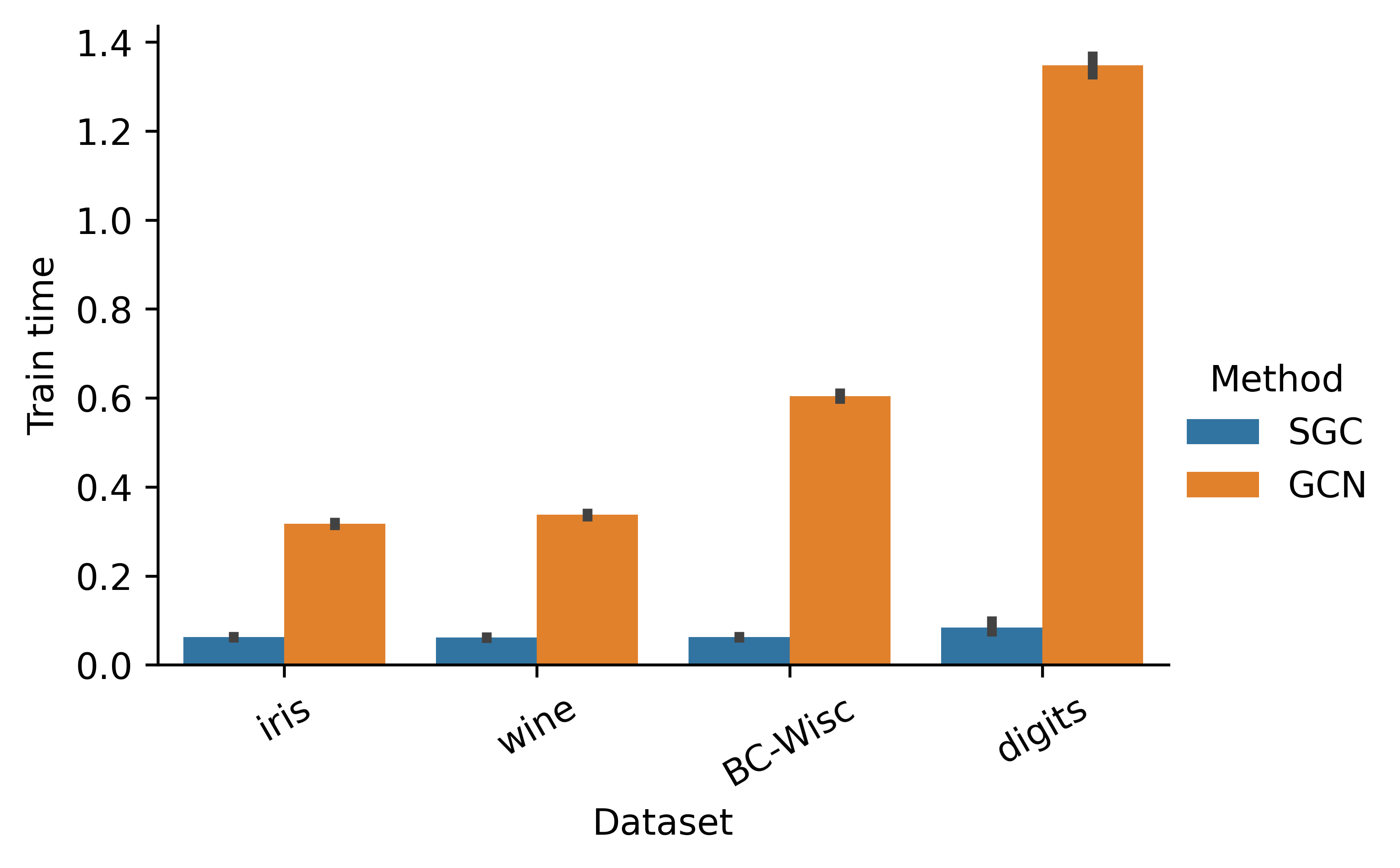}
		\caption{mid-size datasets}
	\end{subfigure}
	\begin{subfigure}[t]{0.32\textwidth}
		\centering\includegraphics[width=\textwidth]{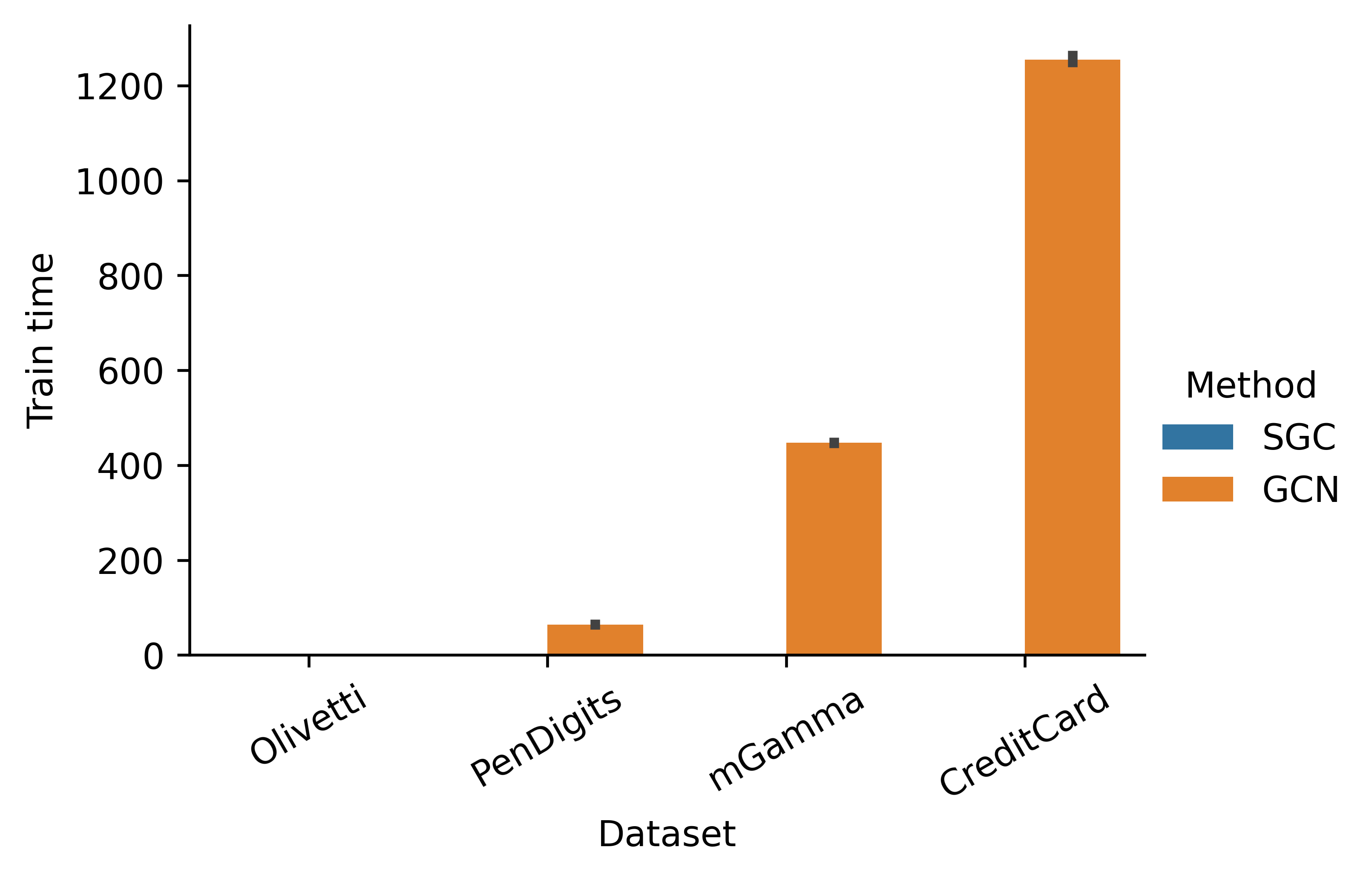}
		\caption{large-size datasets}
	\end{subfigure}
	
	\caption{GCN and SGC training time. (Best viewed in color)}
	\label{Fig:Ex-Time}
\end{figure*}

Another aspect to look at when running GCN and SGC is the training time. The training process in GCN involves feature propagation and nonlinear transformation. While in SGC, the training only involves feature propagation. Figure\ \ref{Fig:Ex-Time} shows the training time for all of the 12 datasets. In case of large-size datasets, the training time for SGC was very small compared to GCN training time. There is a clear advantage for SGC over GCN, especially if we consider that SGC outperforms GCN in most of the datasets. The takeaway from this experiment, we recommend to use SGC because it is faster and most likely will deliver similar or better results than GCN. An exception would be if the user is certain that the dataset contains sparse classes where nonlinearity in GCN can be helpful.

\begin{figure*}
	\centering
	\begin{subfigure}[t]{0.32\textwidth}
		\centering\includegraphics[width=\textwidth]{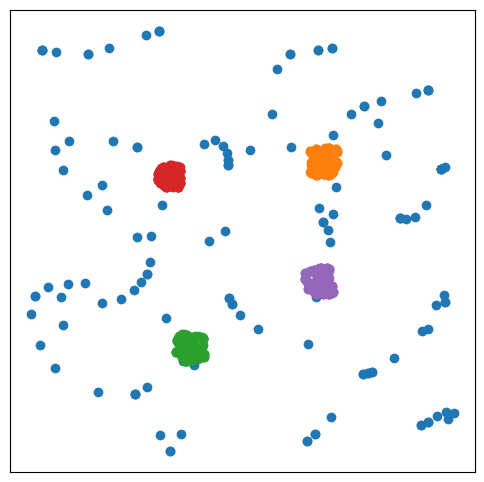}
		\caption{Hidden layers = 2}
	\end{subfigure}
	\begin{subfigure}[t]{0.32\textwidth}
		\centering\includegraphics[width=\textwidth]{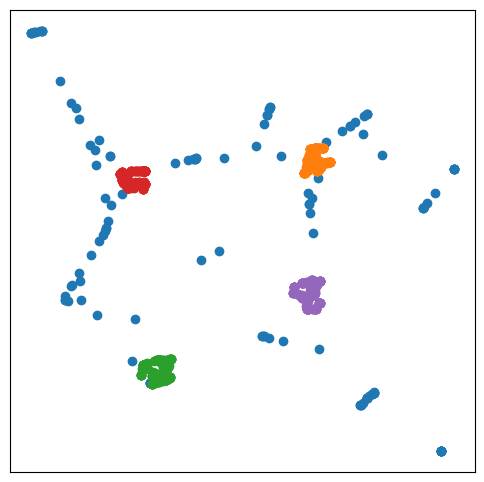}
		\caption{Hidden layers = 50}
	\end{subfigure}
	\begin{subfigure}[t]{0.32\textwidth}
		\centering\includegraphics[width=\textwidth]{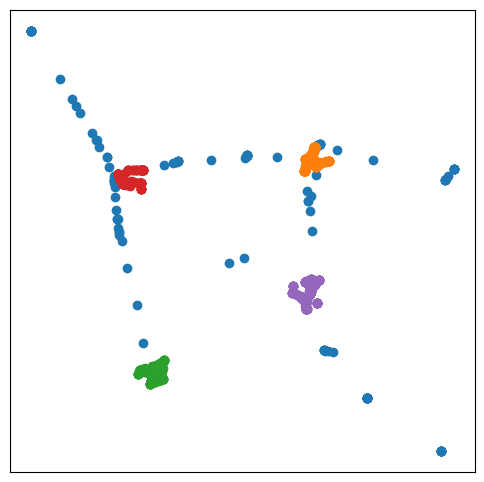}
		\caption{Hidden layers = 200}
	\end{subfigure}
	
	\caption{The effect of graph smoothing in SGC. Smoothing eases the classification task by making features of nodes in the same cluster similar. But oversmoothing makes the features in different clusters indistinguishable (Best viewed in color)}
	\label{Fig:SGC-Smoothing}
\end{figure*}

\begin{figure*}
	\centering
	\includegraphics[width=\textwidth]{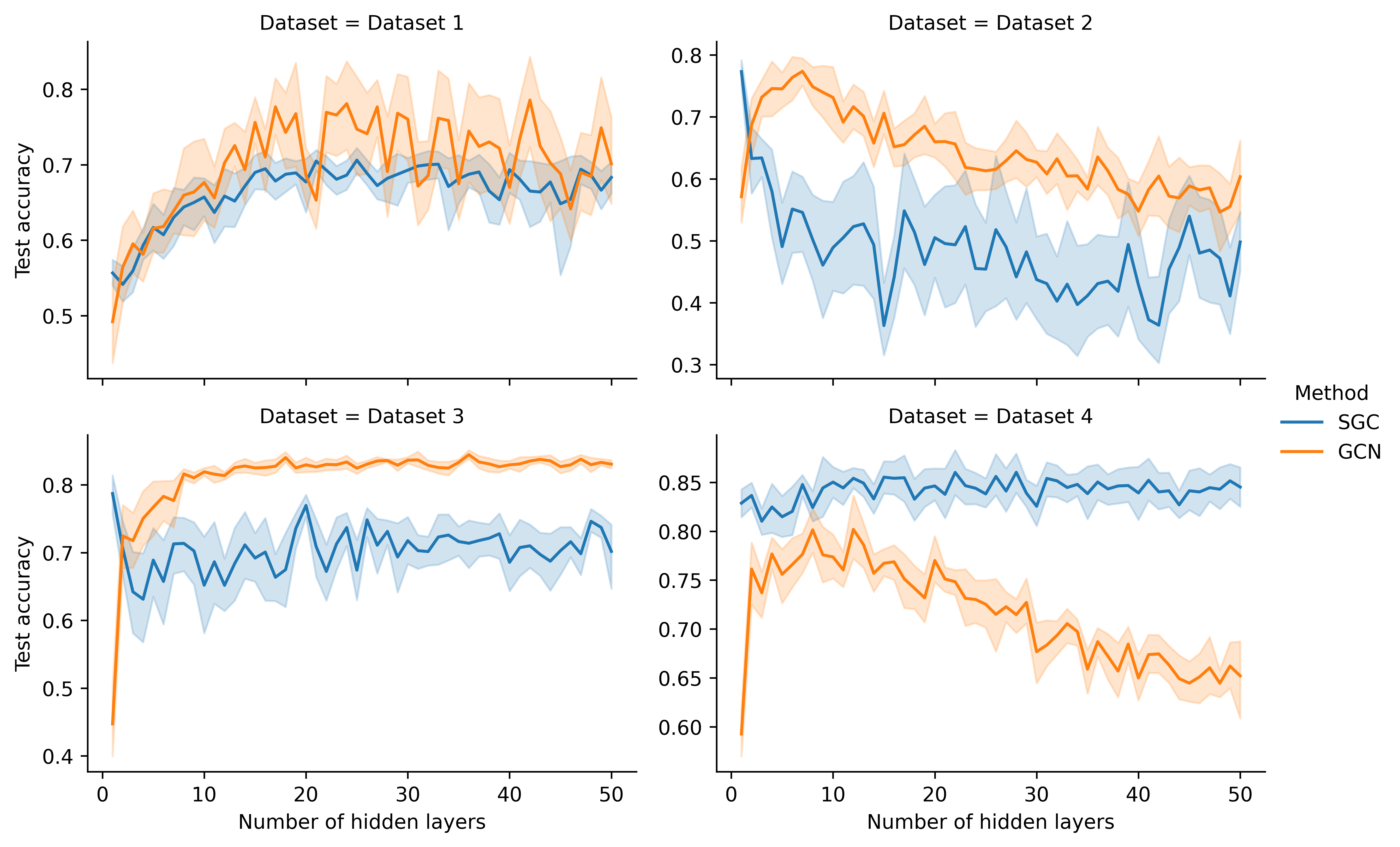}
	\includegraphics[width=\textwidth]{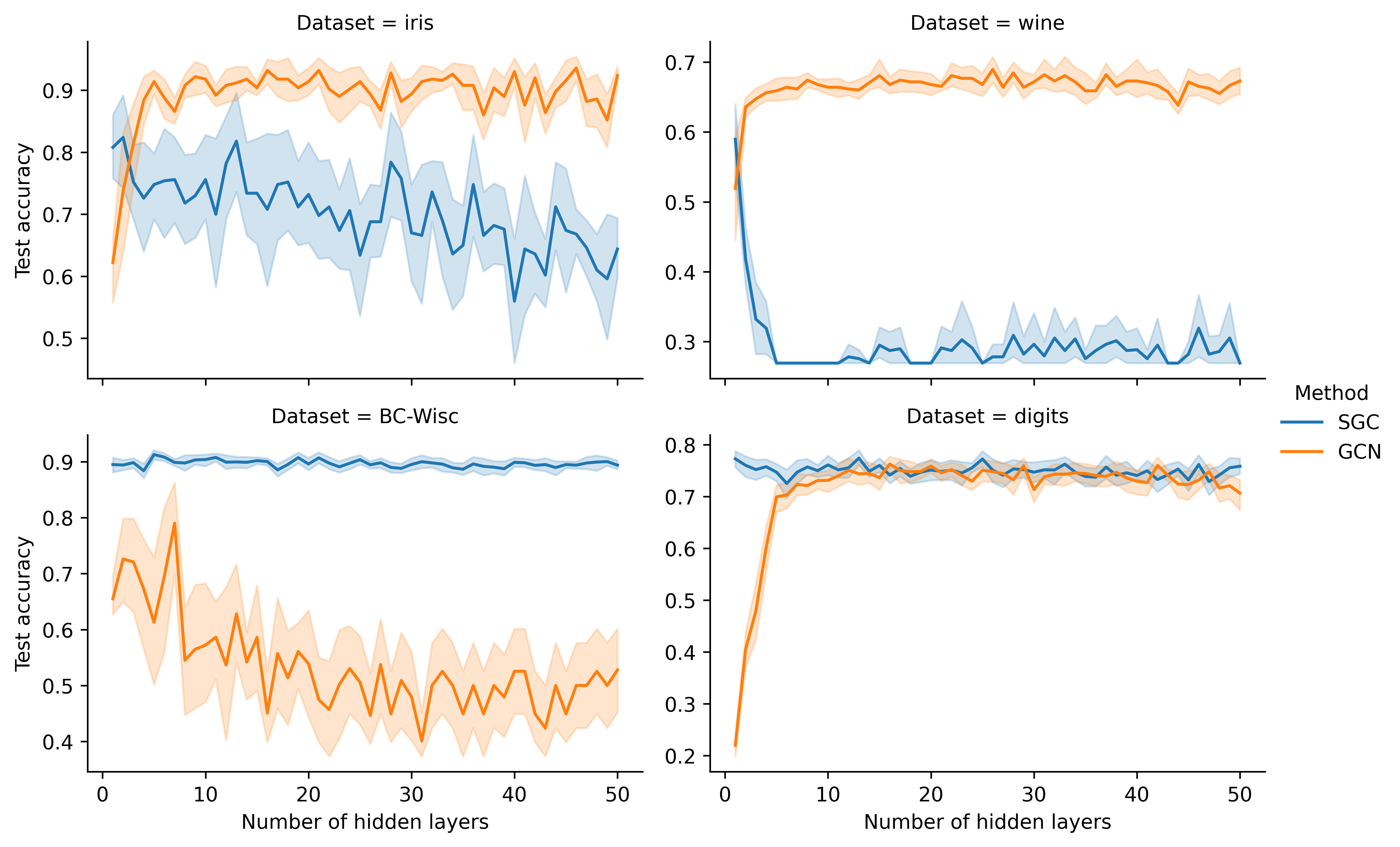}
	\caption{Accuracy scores for SGC and GCN with a varying number of hidden layers; each point on the plot represents an average of 10 runs. (Best viewed in color)}
	\label{Fig:Ex-Smoothing}
\end{figure*}

\subsection{The effect of graph smoothing on GCN and SGC accuracy}
\label{Experiments-Smoothing}
In graph neural network (GNN), each node on the graph $v_i$ is associated with a feature vector $x_i$. The smoothing operation makes the features of nodes in the same cluster similar. This operation eases the classification task. Figure\ \ref{Fig:SGC-Smoothing} shows graph smoothing in SGC.

The number of hidden layers in GCN and SGC controls the graph smoothing. Smoothing in GCN is coupled with nonlinearity transition from one hidden layer to another. While in SGC, smoothing is just a multiplication of the adjacency matrix $A$ by the feature matrix $X$. The risk of oversmoothing is that the features in different clusters became indistinguishable after a number of hidden layers \cite{Yang2020Revisiting}.

In this experiment, we test the effect of graph smoothing on GCN and SGC with our proposed graph construction. Eight datasets were tested, where designed the hidden layers to range from $1$ to $50$ layers for both GCN and SGC. For each network setup we took the test accuracy average for $10$ runs. Out of the eight datasets (see Figure\ \ref{Fig:Ex-Smoothing}), GCN turns to be more resilient than SGC in six datasets. By resilient we mean the ability to deliver better accuracy with an increasing number of layers. These results support the findings presented by \cite{Yang2020Revisiting} where they stated that GCN has the ability to learn ``anti-oversmoothing''.

The two datasets where SGC was better than GCN were \texttt{Dataset 4} and \texttt{BC-Wisc}. This can be explained by the nature of classes in these two datasets, where points in each class share a single mean. In these situations, there is a low risk of oversmoothing. Because with each layer of smoothing, node representations in one class are turning to be the same as the class mean. This is not the case when we have a sparse class. Because with each layer of smoothing node representations in the sparse class are getting mixed with the neighboring classes.

\begin{table}
	\centering
	\includegraphics[width=\textwidth]{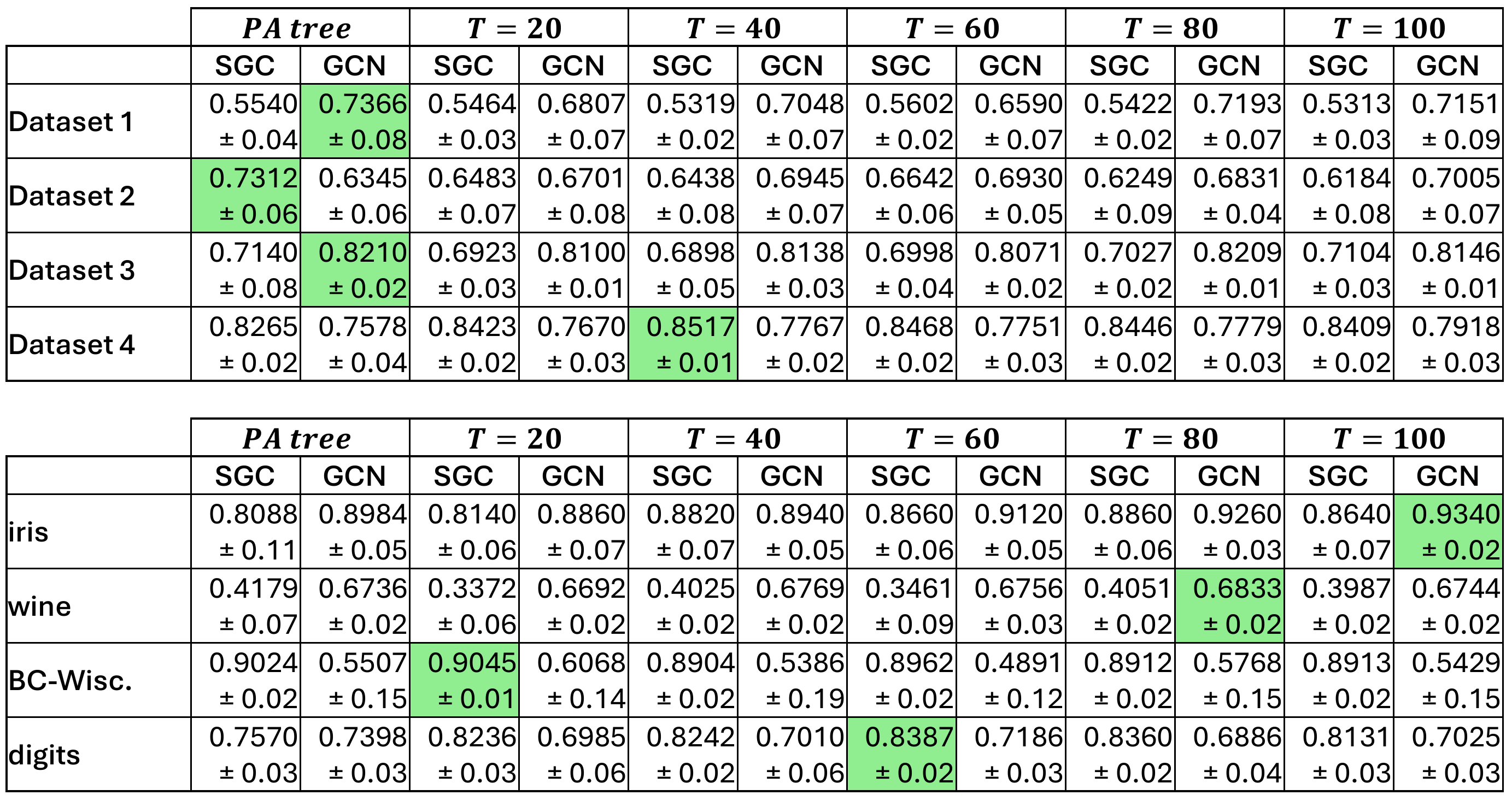}
	\caption{Accuracy scores for SGC and GCN with a varying number of RP-trees; all numbers are averages for 10 runs; the best scores are highlighted in green. (Best viewed in color)}
	\label{Table:Ex-Tree}
\end{table}

\subsection{The effect of the number of trees on GCN and SGC accuracy}
\label{Experiments-Trees}

One factor that affects our graph construction scheme is the reliance on principle axis trees (PA-trees) to construct the graph. In this experiment, we investigate if using another type of Binary Space Partitioning trees (BSP-trees) with an increasing number of trees would improve the test accuracy of GCN and SGC. We used random projection trees (RP-trees) \cite{Dasgupta2008Random,Dasgupta2015Randomized}. In RP-trees, a direction $\vec{r}$ is selected at random, and all points are projected to this random direction. Then the points are split into left and right nodes depending on if they are greater or less than a constant $c$. $c$ is usually set to be the median point along $\vec{r}$. We set the number of trees in a range from $20$ to $100$.

Table\ \ref{Table:Ex-Tree} shows the average test accuracy for $10$ runs. The increasing number of RP-trees will not improve the performance of GCN and SGC. We can explain this by the effect of random directions in RP-trees and the principal component direction is the same. Therefore, the performance of a single PA-tree is similar to the performance of multiple RP-trees. It makes the cost of constructing and storing $100$ RP-trees unjustified.

\subsection{Comparing the constructed graph with a ground-truth adjacency}
\label{ground-truth-adjacency}

There are several datasets with ground truth adjacency matrices. The most used ones in Graph Neural Networks (GNNs) research are \texttt{Citeseer} and \texttt{Cora} \cite{Yang2016Revisiting}. Both datasets are citation network datasets, where the set of nodes $V$ represents the documents and the set of edges $E$ represents the citation links. The features in \texttt{Citeseer} and \texttt{Cora} are bags of words (BoW) representation of documents. \texttt{Citeseer} contains $3,327$ nodes, $9,104$ edges, and $3,703$ features. \texttt{Cora} contains $2,708$ nodes, $10,556$ edges, and $1,433$ features. We used PyG (PyTorch Geometric) \cite{fey2019fast} to download \texttt{Citeseer} and \texttt{Cora} datasets.

For comparison, we used two graph construction methods that are usually used in Graph Neural Networks (GNNs) research: $\epsilon$ graph and $k-$nn graph. The performance of these methods depends heavily on the selection of their hyperparameters. Therefore, we used the recommendations in \cite{Luxburg2007tutorial} to set these hyperparameters. It is advised to set $\epsilon$ equal to the longest edge in the minimum spanning tree (MST). For $k$ in the $k-$nn graph, it is recommended to set it to $log(N)$, where $N$ is the number of instances. For both graphs, we used scikit-learn implementation \cite{scikit-learn} to construct them.

\begin{table}
	\centering
	\includegraphics[width=\textwidth]{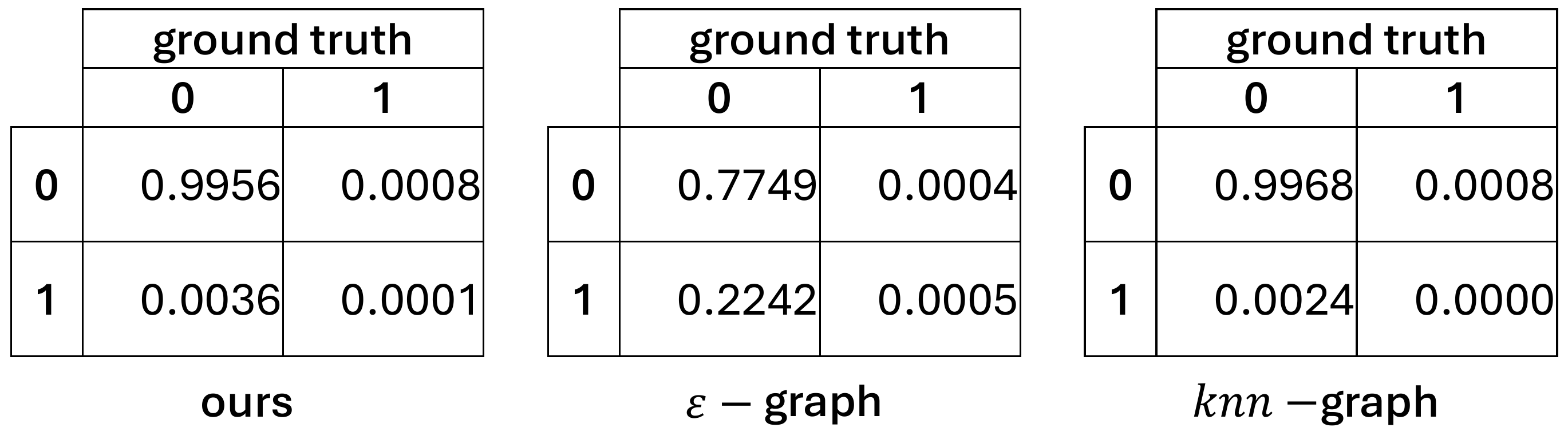}
	\caption{Confusion matrix comparison between \texttt{Citeseer} ground truth adjacency and graph construction methods; 0 indicates no edge and 1 indicates an edge creation.}
	\label{Table:Ex-adj-CiteSeer}
\end{table}

\begin{table}
	\centering
	\includegraphics[width=\textwidth]{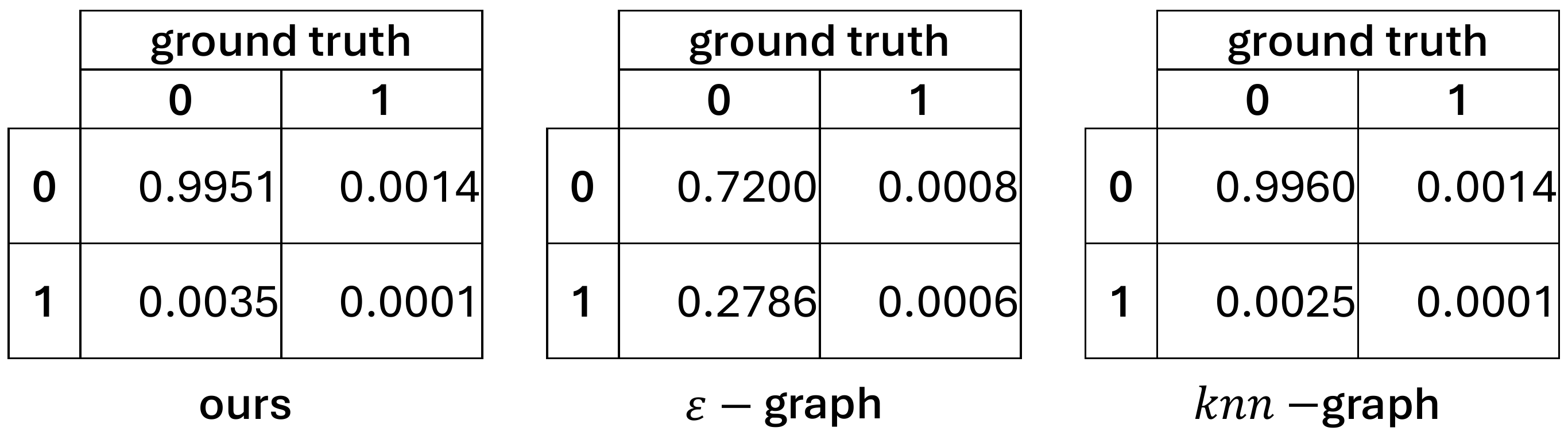}
	\caption{Confusion matrix comparison between \texttt{Cora} ground truth adjacency and graph construction methods; 0 indicates no edge and 1 indicates an edge creation.}
	\label{Table:Ex-adj-CORA}
\end{table}

We compared our proposed method to $\epsilon$ and $k-$nn graphs. With the ground truth adjacency, we constructed a confusion matrix. This confusion matrix has four cases: 1) the edge does not exist in both our graph and the ground truth graph; 2) the edge exists in the ground truth graph but it was missed by our graph; 3) the edge was created by our graph but it does not exist in the ground truth graph; 4) the edge exists in both our graph and the ground truth graph. These confusion matrices are shown in Table \ref{Table:Ex-adj-CiteSeer} and Table \ref{Table:Ex-adj-CORA}.

Our proposed method removed $99.5\%$ of edges compared to \texttt{Citeseer} ground truth graph. That was much higher than the $\epsilon$ graph, which only removed $77.4\%$. The $k$-nn graph removed more edges than our method but it did not get any of the edges created by the ground truth graph. The same observation was noticed in \texttt{Cora} ground truth graph, where the proposed and $k$-nn graphs performed better than the $\epsilon$ graph. The way the ground truth graph was created may be the reason why there is a difference between the ground truth graph and the constructed graphs. The ground truth graph contains citation links, but the features are bags of words. It is not necessarily that two documents with similar bags of words would cite each other. These high-level semantics are missing when we construct a graph only from the features.

\subsection{Comparison with machine learning methods}
\label{machine-learning-methods}

\begin{table}
	\centering
	\includegraphics[width=\textwidth]{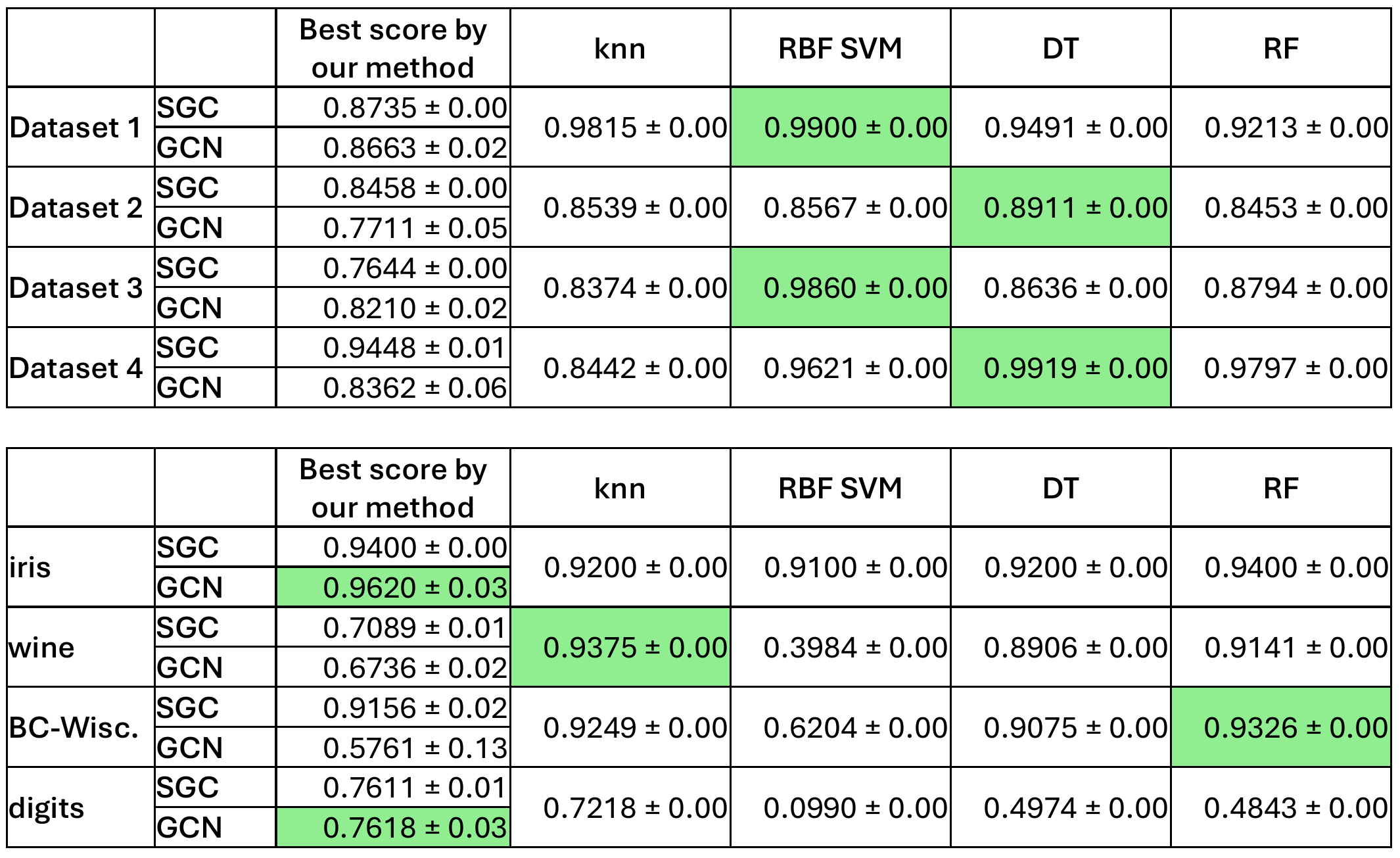}
	\caption{Comparison between the proposed method and other machine learning methods; all scores are averages of 10 runs; the best scores are highlighted in green. (Best viewed in color)}
	\label{Table:Ex-ML-compare}
\end{table}

This experiment was designed to compare our proposed method to well-known machine learning methods that do not require graph construction. We picked four machine learning methods: 1) $k$-nearest neighbor (\texttt{$k$-nn}), 2) support vector machine with radial basis function kernel (\texttt{RBF SVM}), 3) decision tree (\texttt{DT}), 4) random forest (\texttt{RF}). We used scikit-learn implementation \cite{scikit-learn} to run these methods. Our selection for parameters was as follows: $k=5$ for \texttt{$k$-nn}, $\gamma=2$ for \texttt{RBF SVM}, and for \texttt{DT} and \texttt{RF} classifiers we set the maximum depth to 5.

Using eight datasets, we compared our method score to the scores achieved by the other four machine learning methods. These scores are shown in Table \ref{Table:Ex-ML-compare}. Our method achieved the best score on two datasets \texttt{iris} and \texttt{digits}. These scores were achieved by GCN, not SGC. \texttt{RBF SVM} and \texttt{DT} classifiers delivered incosestent performances. They were the best performers on 2-dimensional datasets. However their performance dropped significantly when tested on datasets with higher dimensions. \texttt{$k$-nn} classifier got the highest score only once with \texttt{wine} dataset. This can be explained by the setting of the parameter $k$, which needs to be optimized for each dataset independently.

\subsection{Ablation study}
\label{Ablation-study}

The ablation study is carried out in a way where each component of the proposed method is tested independently. We designed this experiment using four cases. First, is the base case where the graph is constructed by adding edges from the intrinsic graph $A^i$ to the edges found be the PA-tree graph $A^{PA}$, then the edges found in the penalty graph $A^p$ are removed. The base case adjacency is represented as follows $A = (A^{PA}+ A^i)- A^p$. The second case is $A=A^i$, where we only pass the intrinsic graph to the neural net. The third case is represented as $A= A^{PA} - A^p $, where we removed penalty graph edges $A^p$ from PA-tree graph $A^{PA}$. The final case is $A= A^{PA}$, where we ignore all supervised information and only pass the PA-tree graph $ A^{PA}$ to the neural net.

\begin{table}
	\centering
	\includegraphics[width=\textwidth]{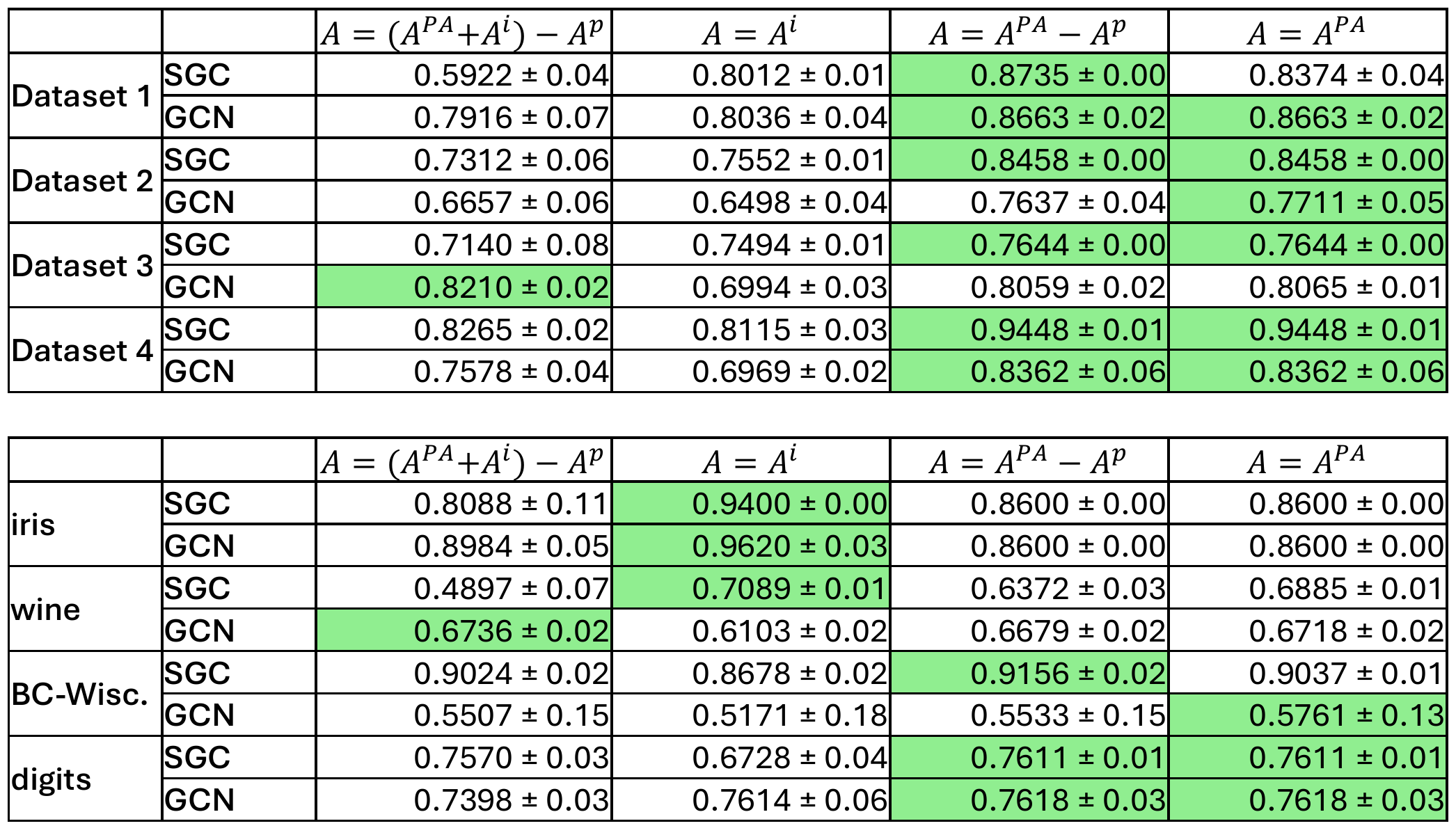}
	\caption{The ablation study results with four cases of graph construction; all scores are averages of 10 runs; the best scores are highlighted in green. (Best viewed in color)}
	\label{Table:Ex-ablation}
\end{table}

All these cases were tested on both SGC and GCN. The results are shown in Table \ref{Table:Ex-ablation}. With 2-dimensional datasets, the intrinsic graph $A^i$ did not achieve the best score. The best scores were achieved mainly by the PA-tree graph $A^{PA}$, either by removing edges from the penalty graph $A^p$ or by passing it as is to the neural net. The intrinsic graph $A^i$ achieved the best score with \texttt{iris} dataset. This could be explained by the percentage of training samples in \texttt{iris} dataset. It has 50 training samples, which is 33\% of the entire dataset. This is the highest percentage of training samples among all tested datasets.

In some cases, we get the same score from $A= A^{PA}$, the PA-tree graph and $A= A^{PA} - A^p$ the PA-tree graph with penalty graph edges removed. This indicates that the penalty graph removed edges that had already been removed by the PA-tree graph.

\subsection{Discussion}
\label{Discussion}

We conducted six experiments to test all aspects of our proposed method. We started by evaluating the accuracy of classification, where SGC and GCN performed similarly. But with the running time advantage, we recommend using SGC instead of GCN because it is faster and most likely will deliver similar performance to GCN.

Two factors that may affect the performance of the proposed method are graph smoothing and the number of trees to construct the graph. We found that SGC performance is vulnerable to a drop in performance if the level of smoothing is not set carefully. For the number of trees factor, we did not find a correlation between increasing the number of trees and performance improvements. Therefore, we recommend using a single PA-tree to construct the graph.

We compared our constructed graph to ground truth graphs that are usually used in the literature. We found that our method constructed a graph that is much closer to ground truth than the one constructed by the $\epsilon$ graph. The $k$-nn graph performed similarly to our method. However, our method has the advantage of not having hyperparameters such as $k$, which might affect the performance of the $k$-nn graph.

We also compared our method to well-known machine learning methods. Some of these methods delivered a strong performance with 2-dimensional data. However, their performance dropped when we tested them with higher dimensions. On the other hand, our method shows resilience to performance drops with higher dimensions.

Our last experiment was the ablation study. In that experiment, we found that the most effective component of our method is the PA-tree graph $A^{PA}$. We constructed this graph based on unsupervised information.

\section{Conclusion}
\label{Conclusion}

Graph Neural Networks (GNNs) have become the go-to option for graph learning among researchers in the machine learning community. GNN eases the computational demands associated with traditional graph learning techniques such as spectral clustering. The adjacency matrix $A$ is crucial for learning in GNN. Despite some efforts to modify the adjacency matrix while training the GNN, these efforts are incompatible with some types of GNNs such as Simple Graph Convolution (SGC).

We present a graph construction scheme that uses unsupervised and supervised information to construct the adjacency matrix $A$. The proposed scheme is independent of GNN training, which makes it compatible with both well-known types of GNNs like Graph Convolutional Networks (GCNs) and Graph Simple Graph Convolution (SGC). We used Principal Axis trees (PA-trees) as a source of unsupervised information to build the adjacency matrix. For supervised information, we used the concepts of penalty and intrinsic graphs from the dimensionality reduction field.

We designed the experiments to examine how GCN and SGC perform with the proposed graph construction in terms of test accuracy and training time. We also examined the factors that could affect the performance (e.g., graph smoothing and the number of BSP-trees). Other experiments compare the proposed method to ground truth adjacency matrices and machine learning methods. We found out that SGC can deliver better or similar test accuracy with far less training time compared to GCN. However, SGC was more vulnerable to the effects of graph smoothing than GCN. We also discovered that the proposed method constructed an adjacency matrix similar to the ground truth matrix. Based on these results, we recommend using SGC with the proposed graph construction because of its speed. But the level of graph smoothing has to be selected carefully.

The proposed method does not use feedback from GCN to optimize the adjacency matrix construction, which might be counted as a drawback. Some methods used a shared loss function for GCN training and adjacency matrix construction. Because we used two different neural net architectures (SGC and GCN), building a feedback loop or shared loss function in these two architectures could be an independent study by itself. Another direction to extend this study could be using other types of Binary Space-Partitioning Trees (BSP-trees) to explore unsupervised information.


\begin{thebibliography}{50}
	\providecommand{\natexlab}[1]{#1}
	\providecommand{\url}[1]{\texttt{#1}}
	\expandafter\ifx\csname urlstyle\endcsname\relax
	  \providecommand{\doi}[1]{doi: #1}\else
	  \providecommand{\doi}{doi: \begingroup \urlstyle{rm}\Url}\fi
	
	\bibitem[Bruna et~al.(2013)Bruna, Zaremba, Szlam, and LeCun]{Bruna2013Spectral}
	Joan Bruna, Wojciech Zaremba, Arthur Szlam, and Yann LeCun.
	\newblock Spectral networks and locally connected networks on graphs, 2013.
	
	\bibitem[Buitinck et~al.(2013)Buitinck, Louppe, Blondel, Pedregosa, Mueller,
	  Grisel, Niculae, Prettenhofer, Gramfort, Grobler, Layton, VanderPlas, Joly,
	  Holt, and Varoquaux]{sklearn_api}
	Lars Buitinck, Gilles Louppe, Mathieu Blondel, Fabian Pedregosa, Andreas
	  Mueller, Olivier Grisel, Vlad Niculae, Peter Prettenhofer, Alexandre
	  Gramfort, Jaques Grobler, Robert Layton, Jake VanderPlas, Arnaud Joly, Brian
	  Holt, and Ga{\"{e}}l Varoquaux.
	\newblock {API} design for machine learning software: experiences from the
	  scikit-learn project.
	\newblock In \emph{ECML PKDD Workshop: Languages for Data Mining and Machine
	  Learning}, pages 108--122, 2013.
	
	\bibitem[Chen et~al.(2023)Chen, Wu, Wang, and Guo]{Chen2023Dual}
	Zhaoliang Chen, Zhihao Wu, Shiping Wang, and Wenzhong Guo.
	\newblock Dual low-rank graph autoencoder for semantic and topological
	  networks.
	\newblock \emph{Proceedings of the AAAI Conference on Artificial Intelligence},
	  37\penalty0 (4):\penalty0 4191--4198, Jun. 2023.
	\newblock \doi{10.1609/aaai.v37i4.25536}.
	
	\bibitem[Dai et~al.(2018)Dai, Li, Tian, Huang, Wang, Zhu, and
	  Song]{Dai2018Adversarial}
	Hanjun Dai, Hui Li, Tian Tian, Xin Huang, Lin Wang, Jun Zhu, and Le~Song.
	\newblock Adversarial attack on graph structured data.
	\newblock In Jennifer Dy and Andreas Krause, editors, \emph{Proceedings of the
	  35th International Conference on Machine Learning}, volume~80 of
	  \emph{Proceedings of Machine Learning Research}, pages 1115--1124. PMLR,
	  10--15 Jul 2018.
	
	\bibitem[Dasgupta and Freund(2008)]{Dasgupta2008Random}
	Sanjoy Dasgupta and Yoav Freund.
	\newblock Random projection trees and low dimensional manifolds.
	\newblock In \emph{Proceedings of the Fortieth Annual ACM Symposium on Theory
	  of Computing}, STOC '08, pages 537--546, New York, NY, USA, 2008. Association
	  for Computing Machinery.
	\newblock ISBN 9781605580470.
	\newblock \doi{https://doi.org/10.1145/1374376.1374452}.
	
	\bibitem[Dasgupta and Sinha(2015)]{Dasgupta2015Randomized}
	Sanjoy Dasgupta and Kaushik Sinha.
	\newblock Randomized partition trees for nearest neighbor search.
	\newblock \emph{Algorithmica}, 72\penalty0 (1):\penalty0 237--263, May 2015.
	\newblock ISSN 0178-4617.
	\newblock \doi{https://doi.org/10.1007/s00453-014-9885-5}.
	
	\bibitem[Defferrard et~al.(2016)Defferrard, Bresson, and
	  Vandergheynst]{Defferrard2016Convolutional}
	Michaël Defferrard, Xavier Bresson, and Pierre Vandergheynst.
	\newblock Convolutional neural networks on graphs with fast localized spectral
	  filtering.
	\newblock 2016.
	\newblock \doi{https://doi.org/10.48550/ARXIV.1606.09375}.
	
	\bibitem[Dua and Graff(2017)]{Dua2019UCI}
	Dheeru Dua and Casey Graff.
	\newblock {UCI} machine learning repository, 2017.
	\newblock URL \url{http://archive.ics.uci.edu/ml}.
	
	\bibitem[Fey and Lenssen(2019)]{fey2019fast}
	Matthias Fey and Jan~Eric Lenssen.
	\newblock Fast graph representation learning with pytorch geometric, 2019.
	
	\bibitem[Franceschi et~al.(2019)Franceschi, Niepert, Pontil, and
	  He]{franceschi2019learning}
	Luca Franceschi, Mathias Niepert, Massimiliano Pontil, and Xiao He.
	\newblock Learning discrete structures for graph neural networks.
	\newblock In \emph{Proceedings of the 36th International Conference on Machine
	  Learning}, 2019.
	
	\bibitem[Fr\"anti and Sieranoja(2018)]{ClusteringDatasets}
	Pasi Fr\"anti and Sami Sieranoja.
	\newblock K-means properties on six clustering benchmark datasets, 2018.
	\newblock URL \url{http://cs.uef.fi/sipu/datasets/}.
	
	\bibitem[Hammond et~al.(2011)Hammond, Vandergheynst, and
	  Gribonval]{Hammond2011Wavelets}
	David~K. Hammond, Pierre Vandergheynst, and Rémi Gribonval.
	\newblock Wavelets on graphs via spectral graph theory.
	\newblock \emph{Applied and Computational Harmonic Analysis}, 30\penalty0
	  (2):\penalty0 129--150, 2011.
	\newblock ISSN 1063-5203.
	\newblock \doi{https://doi.org/10.1016/j.acha.2010.04.005}.
	
	\bibitem[Hart et~al.(2000)Hart, Stork, and Duda]{hart2000pattern}
	Peter~E Hart, David~G Stork, and Richard~O Duda.
	\newblock \emph{Pattern classification}.
	\newblock Wiley Hoboken, 2000.
	
	\bibitem[Henaff et~al.(2015)Henaff, Bruna, and LeCun]{Henaff2015Deep}
	Mikael Henaff, Joan Bruna, and Yann LeCun.
	\newblock Deep convolutional networks on graph-structured data, 2015.
	
	\bibitem[Jia et~al.(2022)Jia, Zhou, Li, Li, and Yin]{JIA2022Data}
	Xibin Jia, Yuhan Zhou, Weiting Li, Jinghua Li, and Baocai Yin.
	\newblock Data-aware relation learning-based graph convolution neural network
	  for facial action unit recognition.
	\newblock \emph{Pattern Recognition Letters}, 155:\penalty0 100--106, 2022.
	\newblock ISSN 0167-8655.
	\newblock \doi{https://doi.org/10.1016/j.patrec.2022.02.010}.
	
	\bibitem[Jin et~al.(2020)Jin, Ma, Liu, Tang, Wang, and Tang]{jin2020graph}
	Wei Jin, Yao Ma, Xiaorui Liu, Xianfeng Tang, Suhang Wang, and Jiliang Tang.
	\newblock Graph structure learning for robust graph neural networks.
	\newblock In \emph{26th ACM SIGKDD International Conference on Knowledge
	  Discovery and Data Mining, KDD 2020}, pages 66--74. Association for Computing
	  Machinery, 2020.
	
	\bibitem[Keivani and Sinha(2021)]{Keivani2021Random}
	Omid Keivani and Kaushik Sinha.
	\newblock Random projection-based auxiliary information can improve tree-based
	  nearest neighbor search.
	\newblock \emph{Information Sciences}, 546:\penalty0 526--542, 2021.
	\newblock \doi{https://doi.org/10.1016/j.ins.2020.08.054}.
	
	\bibitem[Kipf and Welling(2016)]{kipf2016variational}
	Thomas~N Kipf and Max Welling.
	\newblock Variational graph auto-encoders.
	\newblock \emph{NIPS Workshop on Bayesian Deep Learning}, 2016.
	
	\bibitem[Kipf and Welling(2017)]{kipf2017semi}
	Thomas~N. Kipf and Max Welling.
	\newblock Semi-supervised classification with graph convolutional networks.
	\newblock In \emph{International Conference on Learning Representations
	  (ICLR)}, 2017.
	
	\bibitem[Liang et~al.(2021)Liang, Meng, Zhang, Chen, Xu, and
	  Zhou]{LIANG2021dependency}
	Yunlong Liang, Fandong Meng, Jinchao Zhang, Yufeng Chen, Jinan Xu, and Jie
	  Zhou.
	\newblock A dependency syntactic knowledge augmented interactive architecture
	  for end-to-end aspect-based sentiment analysis.
	\newblock \emph{Neurocomputing}, 454:\penalty0 291--302, 2021.
	\newblock ISSN 0925-2312.
	\newblock \doi{https://doi.org/10.1016/j.neucom.2021.05.028}.
	
	\bibitem[Ma et~al.(2023)Ma, Liu, Shah, and Tang]{ma2023homophily}
	Yao Ma, Xiaorui Liu, Neil Shah, and Jiliang Tang.
	\newblock Is homophily a necessity for graph neural networks?, 2023.
	
	\bibitem[McNames(2001)]{McNames2001Fast}
	J.~McNames.
	\newblock A fast nearest-neighbor algorithm based on a principal axis search
	  tree.
	\newblock \emph{IEEE Transactions on Pattern Analysis and Machine
	  Intelligence}, 23\penalty0 (9):\penalty0 964--976, 2001.
	\newblock \doi{https://doi.org/10.1109/34.955110}.
	
	\bibitem[Ng et~al.(2001)Ng, Jordan, and Weiss]{Ng2001Spectral}
	Andrew Ng, Michael Jordan, and Yair Weiss.
	\newblock On spectral clustering: Analysis and an algorithm.
	\newblock In \emph{Advances in Neural Information Processing Systems},
	  volume~14. MIT Press, 2001.
	\newblock URL
	  \url{https://proceedings.neurips.cc/paper/2001/file/801272ee79cfde7fa5960571fee36b9b-Paper.pdf}.
	
	\bibitem[Pedregosa et~al.(2011)Pedregosa, Varoquaux, Gramfort, Michel, Thirion,
	  Grisel, Blondel, Prettenhofer, Weiss, Dubourg, Vanderplas, Passos,
	  Cournapeau, Brucher, Perrot, and Duchesnay]{scikit-learn}
	F.~Pedregosa, G.~Varoquaux, A.~Gramfort, V.~Michel, B.~Thirion, O.~Grisel,
	  M.~Blondel, P.~Prettenhofer, R.~Weiss, V.~Dubourg, J.~Vanderplas, A.~Passos,
	  D.~Cournapeau, M.~Brucher, M.~Perrot, and E.~Duchesnay.
	\newblock Scikit-learn: Machine learning in {P}ython.
	\newblock \emph{Journal of Machine Learning Research}, 12:\penalty0 2825--2830,
	  2011.
	
	\bibitem[Phan et~al.(2022)Phan, Nguyen, and Hwang]{PHAN2022Convolutional}
	Huyen~Trang Phan, Ngoc~Thanh Nguyen, and Dosam Hwang.
	\newblock Convolutional attention neural network over graph structures for
	  improving the performance of aspect-level sentiment analysis.
	\newblock \emph{Information Sciences}, 589:\penalty0 416--439, 2022.
	\newblock ISSN 0020-0255.
	\newblock \doi{https://doi.org/10.1016/j.ins.2021.12.127}.
	
	\bibitem[Qi et~al.(2021)Qi, Zhang, Jia, Mao, Wang, and Song]{QI2021Deep}
	Chao Qi, Jianming Zhang, Hongjie Jia, Qirong Mao, Liangjun Wang, and Heping
	  Song.
	\newblock Deep face clustering using residual graph convolutional network.
	\newblock \emph{Knowledge-Based Systems}, 211:\penalty0 106561, 2021.
	\newblock ISSN 0950-7051.
	\newblock \doi{https://doi.org/10.1016/j.knosys.2020.106561}.
	
	\bibitem[Ram and Gray(2013)]{Ram2013Which}
	Parikshit Ram and Alexander Gray.
	\newblock Which space partitioning tree to use for search?
	\newblock \emph{Advances in Neural Information Processing Systems}, 26, 2013.
	
	\bibitem[Shaham et~al.(2018)Shaham, Stanton, Li, Nadler, Basri, and
	  Kluger]{Shaham2018SpectralNet}
	Uri Shaham, Kelly Stanton, Henry Li, Boaz Nadler, Ronen Basri, and Yuval
	  Kluger.
	\newblock Spectralnet: Spectral clustering using deep neural networks, 2018.
	
	\bibitem[Shi and Malik(1997)]{Shi1997Normalized}
	Jianbo Shi and J.~Malik.
	\newblock Normalized cuts and image segmentation.
	\newblock In \emph{Proceedings of IEEE Computer Society Conference on Computer
	  Vision and Pattern Recognition}, pages 731--737, 1997.
	\newblock \doi{https://doi.org/10.1109/CVPR.1997.609407}.
	
	\bibitem[Shi and Malik(2000)]{Shi2000Normalized}
	Jianbo Shi and J.~Malik.
	\newblock Normalized cuts and image segmentation.
	\newblock \emph{IEEE Transactions on Pattern Analysis and Machine
	  Intelligence}, 22\penalty0 (8):\penalty0 888--905, 2000.
	\newblock \doi{https://doi.org/10.1109/34.868688}.
	
	\bibitem[Shuman et~al.(2013)Shuman, Narang, Frossard, Ortega, and
	  Vandergheynst]{Shuman2013emerging}
	David~I Shuman, Sunil~K. Narang, Pascal Frossard, Antonio Ortega, and Pierre
	  Vandergheynst.
	\newblock The emerging field of signal processing on graphs: Extending
	  high-dimensional data analysis to networks and other irregular domains.
	\newblock \emph{IEEE Signal Processing Magazine}, 30\penalty0 (3):\penalty0
	  83--98, 2013.
	\newblock \doi{https://doi.org/10.1109/MSP.2012.2235192}.
	
	\bibitem[Sproull(1991)]{sproull1991refinements}
	Robert~F Sproull.
	\newblock Refinements to nearest-neighbor searching in k-dimensional trees.
	\newblock \emph{Algorithmica}, 6\penalty0 (1):\penalty0 579--589, 1991.
	
	\bibitem[von Luxburg(2007)]{Luxburg2007tutorial}
	Ulrike von Luxburg.
	\newblock A tutorial on spectral clustering.
	\newblock \emph{Statistics and Computing}, 17, 2007.
	\newblock \doi{https://doi.org/10.1007/s11222-007-9033-z}.
	
	\bibitem[Wang et~al.(2019)Wang, Bian, Liu, and Yan]{Wang2019DC2}
	Ke~Alexander Wang, Xinran Bian, Pan Liu, and Donghui Yan.
	\newblock Dc2: A divide-and-conquer algorithm for large-scale kernel learning
	  with application to clustering.
	\newblock In \emph{2019 IEEE International Conference on Big Data (Big Data)},
	  pages 5603--5610, 2019.
	\newblock \doi{https://doi.org/10.1109/BigData47090.2019.9006565}.
	
	\bibitem[Weiss(1999)]{Weiss1999Segmentation}
	Y.~Weiss.
	\newblock Segmentation using eigenvectors: a unifying view.
	\newblock In \emph{Proceedings of the Seventh IEEE International Conference on
	  Computer Vision}, pages 975--982, 1999.
	\newblock \doi{https://doi.org/10.1109/ICCV.1999.790354}.
	
	\bibitem[Wu et~al.(2019)Wu, Souza, Zhang, Fifty, Yu, and
	  Weinberger]{Wu2019Simplifying}
	Felix Wu, Amauri Souza, Tianyi Zhang, Christopher Fifty, Tao Yu, and Kilian
	  Weinberger.
	\newblock Simplifying graph convolutional networks.
	\newblock In \emph{Proceedings of the 36th International Conference on Machine
	  Learning}, pages 6861--6871. PMLR, 2019.
	
	\bibitem[Wu et~al.(2022)Wu, Shu, Xu, Chang, Chen, and Zheng]{Wu2022Robust}
	Zhebin Wu, Lin Shu, Ziyue Xu, Yaomin Chang, Chuan Chen, and Zibin Zheng.
	\newblock Robust tensor graph convolutional networks via t-svd based graph
	  augmentation.
	\newblock In \emph{Proceedings of the 28th ACM SIGKDD Conference on Knowledge
	  Discovery and Data Mining}, KDD '22, pages 2090--2099. Association for
	  Computing Machinery, 2022.
	\newblock \doi{10.1145/3534678.3539436}.
	
	\bibitem[Wu et~al.(2021)Wu, Pan, Chen, Long, Zhang, and
	  Yu]{Wu2021Comprehensive}
	Zonghan Wu, Shirui Pan, Fengwen Chen, Guodong Long, Chengqi Zhang, and
	  Philip~S. Yu.
	\newblock A comprehensive survey on graph neural networks.
	\newblock \emph{IEEE Transactions on Neural Networks and Learning Systems},
	  32\penalty0 (1):\penalty0 4--24, 2021.
	\newblock \doi{10.1109/TNNLS.2020.2978386}.
	
	\bibitem[Yan et~al.(2018)Yan, Wang, Wang, Wang, and Li]{Yan2018Nearest}
	Donghui Yan, Yingjie Wang, Jin Wang, Honggang Wang, and Zhenpeng Li.
	\newblock K-nearest neighbor search by random projection forests.
	\newblock In \emph{2018 IEEE International Conference on Big Data (Big Data)},
	  pages 4775--4781, 2018.
	\newblock \doi{https://doi.org/10.1109/BigData.2018.8622307}.
	
	\bibitem[Yan et~al.(2019)Yan, Gu, Xu, and Qin]{Yan2019Similarity}
	Donghui Yan, Songxiang Gu, Ying Xu, and Zhiwei Qin.
	\newblock Similarity kernel and clustering via random projection forests, 2019.
	
	\bibitem[Yan et~al.(2021)Yan, Wang, Wang, Wang, and Li]{Yan2021Nearest}
	Donghui Yan, Yingjie Wang, Jin Wang, Honggang Wang, and Zhenpeng Li.
	\newblock K-nearest neighbor search by random projection forests.
	\newblock \emph{IEEE Transactions on Big Data}, 7\penalty0 (1):\penalty0
	  147--157, 2021.
	\newblock \doi{https://doi.org/10.1109/TBDATA.2019.2908178}.
	
	\bibitem[Yan et~al.(2007)Yan, Xu, Zhang, Zhang, Yang, and Lin]{Yan2007Graph}
	Shuicheng Yan, Dong Xu, Benyu Zhang, Hong-jiang Zhang, Qiang Yang, and Stephen
	  Lin.
	\newblock Graph embedding and extensions: A general framework for
	  dimensionality reduction.
	\newblock \emph{IEEE Transactions on Pattern Analysis and Machine
	  Intelligence}, 29\penalty0 (1):\penalty0 40--51, 2007.
	\newblock \doi{https://doi.org/10.1109/TPAMI.2007.250598}.
	
	\bibitem[Yang et~al.(2020)Yang, Wang, Yao, Liu, and
	  Abdelzaher]{Yang2020Revisiting}
	Chaoqi Yang, Ruijie Wang, Shuochao Yao, Shengzhong Liu, and Tarek Abdelzaher.
	\newblock Revisiting over-smoothing in deep {GCNs}, 2020.
	
	\bibitem[Yang et~al.(2016)Yang, Cohen, and Salakhudinov]{Yang2016Revisiting}
	Zhilin Yang, William Cohen, and Ruslan Salakhudinov.
	\newblock Revisiting semi-supervised learning with graph embeddings.
	\newblock In Maria~Florina Balcan and Kilian~Q. Weinberger, editors,
	  \emph{Proceedings of The 33rd International Conference on Machine Learning},
	  volume~48 of \emph{Proceedings of Machine Learning Research}, pages 40--48,
	  New York, New York, USA, 20--22 Jun 2016. PMLR.
	\newblock URL \url{https://proceedings.mlr.press/v48/yanga16.html}.
	
	\bibitem[Ye et~al.(2021)Ye, Sun, Du, Fu, and Xiong]{ye2021coupled}
	Junchen Ye, Leilei Sun, Bowen Du, Yanjie Fu, and Hui Xiong.
	\newblock Coupled layer-wise graph convolution for transportation demand
	  prediction.
	\newblock In \emph{Proceedings of the AAAI Conference on Artificial
	  Intelligence}, volume~35, pages 4617--4625, 2021.
	
	\bibitem[Zelnik-Manor and Perona(2004)]{Zelnik2005Self}
	Lihi Zelnik-Manor and Pietro Perona.
	\newblock Self-tuning spectral clustering.
	\newblock In \emph{Proceedings of the 17th International Conference on Neural
	  Information Processing Systems}, NIPS'04, page 1601–1608, Cambridge, MA,
	  USA, 2004. MIT Press.
	
	\bibitem[Zhang et~al.(2022)Zhang, Zhu, Hou, and Yin]{Zhang2022Graph}
	Shi-Xue Zhang, Xiaobin Zhu, Jie-Bo Hou, and Xu-Cheng Yin.
	\newblock Graph fusion network for multi-oriented object detection.
	\newblock \emph{Applied Intelligence}, 2022.
	\newblock \doi{https://doi.org/10.1007/s10489-022-03396-5}.
	
	\bibitem[Zhao and Akoglu(2019)]{Zhao2019PairNorm}
	Lingxiao Zhao and Leman Akoglu.
	\newblock Pairnorm: Tackling oversmoothing in {GNNs}, 2019.
	
	\bibitem[Zhong et~al.(2023)Zhong, Yang, Chen, and Wang]{Zhong2023Contrastive}
	Luying Zhong, Jinbin Yang, Zhaoliang Chen, and Shiping Wang.
	\newblock Contrastive graph convolutional networks with generative adjacency
	  matrix.
	\newblock \emph{IEEE Transactions on Signal Processing}, 71:\penalty0 772--785,
	  2023.
	\newblock \doi{10.1109/TSP.2023.3254888}.
	
	\bibitem[Zhou et~al.(2020)Zhou, Huang, Hu, and He]{ZHOU2020Modeling}
	Jie Zhou, Jimmy~Xiangji Huang, Qinmin~Vivian Hu, and Liang He.
	\newblock {SK-GCN}: Modeling syntax and knowledge via graph convolutional
	  network for aspect-level sentiment classification.
	\newblock \emph{Knowledge-Based Systems}, 205:\penalty0 106292, 2020.
	\newblock ISSN 0950-7051.
	\newblock \doi{https://doi.org/10.1016/j.knosys.2020.106292}.
	
	\end{thebibliography}

\end{document}